\documentclass{article}

\PassOptionsToPackage{numbers, compress}{natbib}

\usepackage[final]{neurips_2021}

\usepackage[utf8]{inputenc} 
\usepackage[T1]{fontenc}    
\usepackage{hyperref}       
\usepackage{url}            
\usepackage{booktabs}       
\usepackage{nicefrac}       
\usepackage{microtype}      

\usepackage{times}
\usepackage{epsfig}

\usepackage{url}

\usepackage{color}
\usepackage[dvipsnames]{xcolor}
\usepackage{bbm}
\usepackage{graphicx}
\usepackage{amssymb}
\usepackage{amsfonts}
\usepackage{amsthm}
\usepackage{multirow}
\usepackage{enumitem}
\usepackage{framed}
\usepackage{amsmath}
\usepackage{mathtools}
\usepackage{subfigure}
\usepackage{caption}
\usepackage{footmisc}
\usepackage{bm}
\usepackage{diagbox}

\title{Visualizing the Emergence of Intermediate Visual Patterns in DNNs}

\author{%
  Mingjie Li\\
  Shanghai Jiao Tong University\\
  \texttt{limingjie0608@sjtu.edu.cn} \\
   \And
   Shaobo Wang\\
   Harbin Institute of Technology \\
   \texttt{181110315@stu.hit.edu.cn} \\
   \And
   Quanshi Zhang\thanks{Corresponding author. This work was done under the supervison of Dr. Quanshi Zhang. He is with the John Hopcroft Center and the MoE Key Lab of Artificial Intelligence, AI Institute, at the Shanghai Jiao Tong University, China.}\\
   Shanghai Jiao Tong University\\
   \texttt{zqs1022@sjtu.edu.cn}
}

\begin{document}

\maketitle

\begin{abstract}
This paper proposes a method to visualize the discrimination power of intermediate-layer visual patterns encoded by a DNN.
Specifically, we visualize (1) how the DNN gradually learns regional visual patterns in each intermediate layer during the training process, and (2) the effects of the DNN using non-discriminative patterns in low layers to construct disciminative patterns in middle/high layers through the forward propagation.
Based on our visualization method, we can quantify knowledge points (\emph{i.e.} the number of discriminative visual patterns) learned by the DNN to evaluate the representation capacity of the DNN.
Furthermore, this method also provides new insights into signal-processing behaviors of existing deep-learning techniques, such as adversarial attacks and knowledge distillation.
\end{abstract}

\section{Introduction}

Deep neural networks (DNNs) have achieved superior performance in various tasks, but the black-box nature of DNNs makes it difficult for people to understand its internal behavior.
Visualization methods are usually considered as the most direct way to understand the DNN.
Recently, several attempts have been made to visualize the DNN from different aspects, \emph{e.g.} illustrating the visual appearance that maximizes the prediction score of a given category~\cite{simonyan2013deep,yosinski2015understanding,google2015inceptionism}, inverting intermediate-layer features to network inputs~\cite{dosovitskiy2016inverting}, extracting receptive fields of neural activations~\cite{zhou2014object}, estimating saliency/importance/attribution maps~\cite{zhou2016learning,selvaraju2017grad,zintgraf2017visualizing,lunberg2017unified}, visualizing the sample distribution, such as PCA~\cite{pearson1901liii}, t-SNE~\cite{van2008visualizing}, etc.

In spite of above explanations of the DNN, there is still a large gap between visual explanations of the patterns in the DNN and the theoretical analysis of the DNN's discrimination power.
In other words, visualization results usually cannot reflect the discrimination power of features in the DNN.

Therefore, instead of simply visualizing the entire sample, we divide intermediate-layer features into feature components, each of which represents a specific image region.
We visualize the discrimination power of these feature components, and we consider discriminative feature components as knowledge points learned by the DNN.
Based on above methods, we can diagnose the feature representation of a pre-trained DNN from the following perspectives.

\begin{figure}[t]
    \centering
    \includegraphics[width=\linewidth]{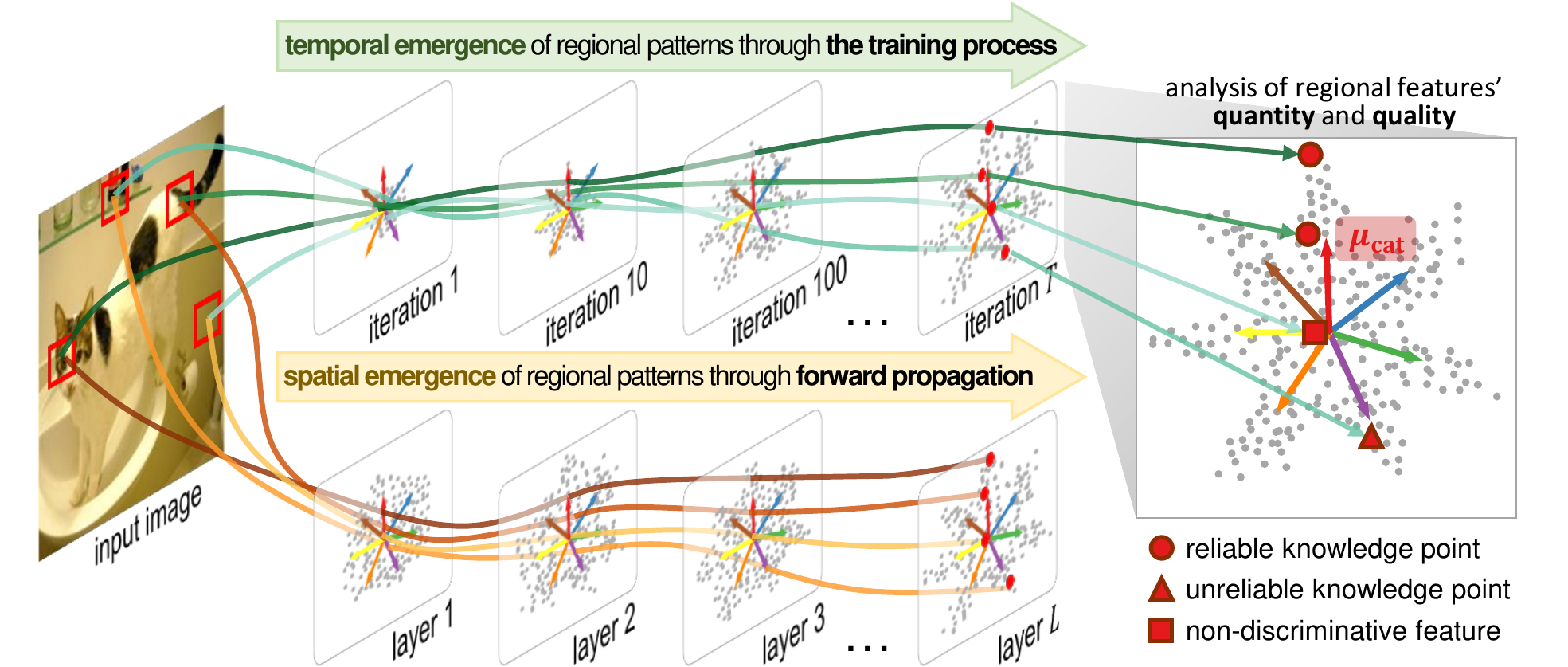}
    \caption{{\small Diagrammatic sketch for the emergence of visual patterns encoded by the DNN in a temporal-spatial manner. The visualization result enables people to analyze the quantity and quality of intermediate features.}}
    \vspace{-1em}
    \label{fig:figure1}
\end{figure}

$\bullet$ We visualize the emergence of intermediate visual patterns in a temporal-spatial manner and evaluate their discrimination power.
(1) We visualize how the discrimination power of each individual visual pattern increases during the learning process. (2) We illustrate effects of using non-discriminative patterns in low layers to gradually construct discriminative patterns in high layers during the forward propagation.
As Figure~\ref{fig:figure1} shows, the regional feature of the cat head emerges as a discriminative pattern for the cat category, while wall features are non-discriminative.

$\bullet$ Based on the the visualization result, we can further measure the quantity and quality of intermediate patterns encoded by a DNN.
In Figure~\ref{fig:figure1}, we count knowledge points encoded in a DNN as regional patterns with strong discrimination power, and further evaluate whether each knowledge point is reliable for classification.
This provides a new perspective to analyze the DNN.


A distinct contribution of this study is to bridge the empirical visualization and the quantitative analysis of a DNN's discrimination power.
In comparison, \citet{kim2018interpretability} used concept activation vectors to model the relationship between visual features and manually annotated semantic concepts.
\citet{cheng2020explaining} quantified the number of visual concepts encoded by the DNN.
However, these two methods cannot reflect the discrimination power of regional visual concepts.
On the other hand, some researchers derived mathematical bounds on the representational power of a DNN~\cite{zhang2016understanding,du2018many} under certain assumptions of the network architecture.
To this end, we believe that bridging regional patterns and a DNN's discrimination power is a more convincing and more intuitive way to reveal the internal behavior of a DNN than mathematical bounds under certain assumptions.

Besides, our method provides insightful understanding towards existing deep-learning techniques, such as adversarial attacks and knowledge distillation.
(1) For adversarial attacks, we discover that adversarial attacks mainly affect unreliable regional features in high layers of the DNN.
The visualization result also enables us to categorize attacking behaviors of all image regions into four types.
(2) For knowledge distillation, we discover that the student DNN usually encodes less reliable knowledge points, compared with the teacher DNN.
Although knowledge distillation is able to force the student DNN to mimic features of a specific layer in the teacher DNN, there is still a big difference of features in other layers between the student DNN and the teacher DNN.

\textbf{Contributions} of this paper can be summarized as follows. (1) We propose a method to visualize the discrimination power of intermediate-layer features in the DNN, and illustrate the emergence of intermediate visual patterns in a temporal-spatial manner. (2) Based on the visualization result, we quantify knowledge points encoded by a DNN. (3) The proposed method also provides new insights into existing deep-learning techniques, such as the adversarial attack and the knowledge distillation.

\section{Related work}

\textbf{Visual explanations for DNNs.}
Visualization of DNNs is the most direct way to explain visual patterns encoded in the DNN.
Some studies reconstructed the network input based on a given intermediate-layer feature~\cite{mahendran2015understanding,simonyan2017deep,dosovitskiy2016inverting}, while others aimed to generate an input image that cause high activations in a given feature map~\cite{zeiler2014visualizing}.
\citet{zhou2014object} extracted the actual image-resolution receptive field of neural activations in DNNs.
Recently, \citet{goh2021multimodal} visualized the multi-modal neurons in the CLIP models~\cite{radford2021learning}.
\citet{kim2018interpretability} proposed TCAV to represent manually annotated semantic concepts.
Another type of researches diagnosed and visualized the pixel-wise saliency/importance/attribution on real input images~\cite{ribeiro2016should,lunberg2017unified,kindermans2017learning,zintgraf2017visualizing,fong2017interpretable,zhou2016learning,chattopadhay2018grad}.
What is more, the distribution of intermediate-layer features could be visualized via dimensionality reduction methods, \emph{e.g.} PCA~\cite{pearson1901liii}, LLE~\cite{roweis2000nonlinear}, MDS~\cite{cox2008multidimensional}, ISOMAP~\cite{tenenbaum2000global}, t-SNE~\cite{van2008visualizing}, etc.
Recently, \citet{law2019dimensionality} visualized the feature of each sample in a low-dimensional space by exploiting probability scores of the DNN, while \citet{li2020visualizing} achieved this by assigning each category in the task with a main direction.
Sophisticated interfaces have been built up to visualize the architecture and the knowledge of DNNs~\cite{harley2015interactive,wang2020cnn,tenney2020language}.

Instead of visualizing visual appearance or merely visualizing the distribution of sample features, our method visualizes the emergence of intermediate visual patterns in a temporal-spatial manner, which provides a new perspective to explain DNNs.

\textbf{Theoretical analysis of the representation capacity of DNNs.}
Formulating and evaluating the discrimination power of DNNs is another direction to explain DNNs.
The information-bottleneck theory~\cite{wolchover2017new,shwartz2017opening} provides a generic metric to quantify information encoded in DNNs, which is further exploited to evaluate the representation capacity of DNNs~\cite{goldfeld2019estimating,xu2017information}. \citet{achille2018information} further used the information-bottleneck to improve the feature representation.
Furthermore, several metrics were also proposed to analyze the robustness or generalization capacity of DNNs, such as the CLEVER score~\cite{weng2018evaluating}, the stiffiness~\cite{fort2019stiffness}, the sensitivity metrics~\cite{novak2018sensitivity}, etc.
\citet{zhang2019all} studied the role of different layers towards the generalization of deep models.
Based on this, module criticality~\cite{chatterji2019intriguing} was proposed to analyze a DNN's generalization power.
\citet{cheng2020explaining} quantified visual concepts in input images from the perspective of pixel-wise entropies.
\citet{liang2019knowledge} diagnosed feature representation between different DNNs via knowledge consistency.
Some studies also theoretically proved the generalization bound for two-layer neural networks~\cite{zhang2016understanding,du2018many,neyshabur2018towards} and deep CNNs~\cite{long2019generalization,li2018tighter}.

In fact, our research group led by Dr. Quanshi Zhang have proposed game-theoretic interactions as a new perspective to explain the representation capacity of trained DNNs.
The interactions have been used to explain the hierarchical structure of information processing in DNNs~\cite{zhang2021nlptree,zhang2021multivariate}, generalization power~\cite{zhang2020dropout}, the complexity~\cite{cheng2021concepts} and the aesthetic level~\cite{cheng2021aesthetic} of the encoded visual concepts, adversarial robustness~\cite{ren2021advsarial,wang2021adversarial}, and adversarial transferability~\cite{wang2020transfer} of DNNs. The interaction was also used to learn baseline values of Shapley values ~\cite{ren2021baseline}.

However, there is still a lack of connections between the explanations of visual concepts and the analysis of a DNN's discrimination power.
To this end, our method enables people to use the discrimination power of local regions to explain the overall discrimination power of the entire DNN.

\section{Algorithm}

Given a pre-trained DNN, we propose an algorithm to project the feature of an entire sample and features corresponding to different image regions into a low-dimensional space, in order to visualize (1) how the discrimination power of a sample feature increases during the learning process, (2) effects of the DNN constructing discriminative features in high layers using non-discriminative features in low layers, and (3) how discriminative regional features gradually emerge during the training process.
The visualization result reflects whether the sample feature and each regional feature are biased to incorrect categories, which regional features play a crucial role in classification, whether a regional feature is shared by multiple categories, etc.
Furthermore, the visualization enables people to quantify knowledge points encoded in the DNN.
Such analysis can help people explain existing deep-learning techniques, \emph{e.g.} adversarial attacks and knowledge distillation.

\subsection{Radial distribution to analyze the discrimination power of features}
\label{sec:algo-radial}

\textbf{Preliminaries: the vMF distribution.}
Given a pre-trained DNN and an input image {\small$x\in\mathbb{R}^n$}, let us consider the output feature of a specific layer, denoted by {\small$f\in\mathbb{R}^d$}.
To analyze the discimination power of {\small$f$}, previous literature usually considered the feature to follow a radial distribution of massive pseudo-categories (much more than the real category number)~\cite{wen2016discriminative, wang2017normface, law2019dimensionality}.
As Figure \ref{fig:figure1}(right) shows, each category/pseudo-category {\small$c$} has a mean direction {\small$\mu_c$ ($c=1,...,C$)} in the feature space.
The significance of classifying {\small$x$} towards category {\small$c$} is measured by the projection {\small$f^\top\mu_c$}, and {\small$\cos(f,\mu_c)$} indicates the similarity between {\small$f$} and category {\small$c$}.
For example, a typical case is the softmax operation, {\small$p(y|f)=\text{Softmax}(Wf)$}, {\small$W\in\mathbb{R}^{C\times d}$}. The {\small$c$}-th row of {\small$W$} indicates the direction corresponding to the {\small$c$}-th category.

To this end, the von Mises-Fisher (vMF) mixture model~\cite{banerjee2005clusteringvmf,hasnat2017mises} was proposed to model the radial distribution, where the {\small$c$}-th mixture component assumes that the likelihood of each feature {\small$f$} belonging to category {\small$c\in\{1,...,C\}$} follows a vMF distribution.\vspace{-3pt}
\begin{equation}
\begin{small}
    p(f)={\sum}_c p(y=c)p_{\text{vMF}}(f|y=c)
    \text{,\quad}
    p_{\text{vMF}}(f|y=c)=C_d(\kappa_c)\cdot\exp[\kappa_c\cdot\cos(\mu_c,f)],
    \label{eq:vmf-pdf}
\end{small}
\end{equation}
where {\small$C_d(\kappa_c)=\frac{\kappa_c^{d/2-1}}{(2\pi)^{d/2}I_{d/2-1}(\kappa_c)}$} is the normalization constant.
Actually, the vMF distribution can be considered as a spherical analogue to the Gaussian distribution on the unit sphere.
{\small$\mu_c\in\mathbb{R}^d$} measures the mean direction of category {\small$c$}.
The increase of {\small$\kappa_c\geq0$} decreases the variance of {\small$f$}'s orientation \emph{w.r.t.} the mean direction {\small$\mu_c$}.
Please see the supplementary material for more details.

\textbf{Radial distribution with noise.}
The vMF distribution assumes that {\small$f$} is a clean feature without noise, which makes the inference of {\small$f$} purely based on its orientation and independent of the strength {\small$\Vert f\Vert_2$}.
However, in real applications, {\small$f$} usually contains the meaningful and clean feature {\small$f^\star$} and the meaningless noise {\small$\epsilon\sim\mathcal{N}(0,\sigma^2I_{d})$}, \emph{i.e.} {\small$f=f^\star+\epsilon$}.
The existence of the noise decreases the classification confidence if the strength of {\small$f$} is low.
Therefore, we have\vspace{-5pt}
\begin{small}
\begin{equation}
    p(f|y=c)=\int p(\epsilon)\cdot p_{\text{vMF}}\Big({f^\star}=f-\epsilon|y=c\Big)\ d\epsilon.
    \label{eq:noise-probability}
\end{equation}
\end{small}
For simplicity, we can assume that all features {\small$f$} of a specific strength {\small$l$} follow a vMF distribution, because they have similar vulnerabilities to noises.
Specifically, let {\small$f=[o,l]$}, where {\small$l=\Vert f\Vert_2$} and {\small$o=f/l$} represent the strength and orientation of {\small$f$}.
Then, the likelihood of {\small$f$} belonging to category {\small$c$} is given as follows (proof in the supplementary material).\vspace{-3pt}
\begin{small}
\begin{equation}
    p(f=[o,l]|y=c)=p(l|y=c)\cdot p_{\text{vMF}}(o|y=c,l)=p(l|y=c)\cdot p_{\text{vMF}}(o|\mu_c,\kappa(l)),
    \label{eq:extended-vmf}
\end{equation}
\end{small}
where {\small$\kappa(l)$} increases along with {\small$l=\Vert f\Vert_2$}, and {\small$p(l|y=c)$} is the prior distribution of {\small$\Vert f\Vert_2$} for category {\small$c$}.
The variance parameter {\small$\kappa(l)$} is determined based on statistics of all features of the same strength {\small$l$}.
We can prove the classification result {\small$p(y\!=\!c|f)$} is confident when the feature {\small$f$} has a large strength.

\subsection{Visualization of the sample-wise discrimination power}
\label{sec:algo-sample}

In this section, we visualize the discrimination power of the feature {\small$f\in\mathbb{R}^d$} of each entire sample {\small$x\in X$}.
The visualization is supposed to illustrate the classification confidence of each feature towards different categories.
Therefore, the goal is to learn a linear transformation to project {\small$f$} into a low-dimensional space, \emph{i.e.} {\small$g\!=\!M f\!\in\!\mathbb{R}^{d'}$ $(d'\!\ll\!d)$}, which ensures that the similarity between each sample feature {\small$f$} and different categories is preserved.
The basic idea of learning the linear transformation {\small$M$} is to use the projected feature {\small$g$} for classification, and to force the classification based on {\small$g$} to mimic the classification based on the original feature {\small$f$}.
Let {\small$y\in Y=\{1,...,C\}$} denote a category.
Thus, the objective is to minimize the KL divergence between the classification probability of the DNN {\small$P(y|x)$} and the classification probability {\small$Q_{M}(y|x)$} based on the projected feature {\small$g$}.\vspace{-4pt}
\begin{small}
\begin{equation}
    \min_{M}\ KL\left[P(Y|X)\Vert Q_{M}(Y|X)\right]
    \Rightarrow
    \min_{M}\
    \mathbb{E}_x\left[{\sum}_{y}P(y|x)\log\frac{P(y|x)}{Q_{M}(y|x)}\right],
    \label{eq:s-emb-goal-func}
\end{equation}
\end{small}
where {\small$P(y|x)$} is usually computed using a softmax operation.
{\small$Q_{M}(y|x)$} is computed by assuming the distribution {\small$p(g)$} as a mixture model, where each mixture component {\small$p(g|y)$} follows a revised vMF distribution in Eq. (\ref{eq:extended-vmf}).
Let {\small$g=[l_g,o_g]$}, where {\small$l_g=\Vert g\Vert_2$} and {\small$o_g=g/l_g$} denote the strength and orientation of {\small$g$}.
Then, we have\vspace{-4pt}
\begin{small}
\begin{equation}
    p\left(g\right)
    ={\sum}_{y} \pi_y\cdot p\left(l_g|y\right)\cdot p_{\text{vMF}}\left(o_g|\mu_y,\kappa(l_g)\right),
\end{equation}
\end{small}
where {\small$\pi_y$} denotes the prior of the {\small$y$}-th category.
We assume the prior of {\small$g$}'s strength is independent with the category, \emph{i.e.} {\small$p(l_g|y)=p(l_g)$}.
Then, {\small$Q_{M}(y|x)$} can be measured by the posterior probability {\small$p(y|g)$} in the mixture model, \emph{i.e.}\vspace{-4pt}
\begin{small}
\begin{equation}
    Q_M(y|x)
    =\frac{p(y)\cdot p(g|y)}{p(g)}
    =\frac{\pi_{y}\cdot p_{\text{vMF}}(o_g|\mu_{y},\kappa(l_g))}{\sum_{y'}\pi_{y'}\cdot p_{\text{vMF}}(o_g|\mu_{y'},\kappa(l_g))}.
    \label{eq:posterior}
\end{equation}
\end{small}
The training of the sample-level visualization alternates between the following two steps. (i) Given the current linear transformation {\small$M$}, we update the mixture-model parameters {\small$\{\pi,\mu\}=\{\pi_y,\mu_y\}_{y\in Y}$} via the maximium likelihood estimation (MLE) {\small$\max_{\{\pi,\mu\}}\prod_{g}p(g)$}. (ii) Given the current state of {\small$\{\pi,\mu\}$}, we update {\small$M$} to minimize the KL divergence {\small$KL(P(Y|X)\Vert Q_M(Y|X))$} in Eq. (\ref{eq:s-emb-goal-func}).
The supplementary material provides more discussions and derivations about the learning process.

\subsection{Visualization of the regional discrimination power}

In this section, we visualize the discrimination power of features extracted from different regions in each input sample.
Let {\small$f\!\!\in\!\mathbb{R}^{K\!\times\! H\!\times\! W}$} be an intermediate-layer feature of a sample {\small$x$}, which is composed of {\small$HW$} regional features for {\small$HW$} positions. Each {\small$r$}-th regional feature is a {\small$K$}-dimensional vector, and is supposed to mainly describe the {\small$r$}-th region in the input, corresponding to the receptive field (region) of {\small$f^{(r)}$}.
In fact, the actual receptive field of {\small$f^{(r)}$} is much smaller than the theoretical receptive field~\cite{zhou2014object}.
The discrimination power of each regional feature {\small$f^{(r)}$} is analyzed in terms of both the \textit{importance} and \textit{reliability}, \emph{i.e.} (1) whether {\small$f^{(r)}$} has a significant impact on the classification, and (2) whether {\small$f^{(r)}$} pushes the classification towards the ground-truth category without significant bias.

The visualization of regional discrimination power needs two overcome to challenges. First, we need to formulate and estimate specific importance of different regions for inference during the learning of visualization. Second, the visualization of regional discrimination power is supposed to be aligned with the coordinate system for sample-wise discrimination power.

The goal of the visualization is to project regional features into a low-dimensional space via a linear transformation {\small$\Lambda$}, \emph{i.e.} {\small$h^{(r)}=\Lambda f^{(r)}\in\mathbb{R}^{d'}$ $(d'\!\ll\!K)$}.
Each projected regional feature {\small$h^{(r)}$} is supposed to reflect the \textit{importance} and \textit{reliability} of the original feature {\small$f^{(r)}$}.
The strength {\small$\Vert h^{(r)}\Vert_2$} reflects the importance, and the orientation of {\small$h^{(r)}$} represents the reliability of classification towards different categories.
Just like t-SNE~\cite{van2008visualizing}, we use the projected features {\small$\bm{h}=\{h^{(1)},...,h^{(HW)}\}$} to infer the similarity between samples.
In this study, the distinct idea of learning {\small$\Lambda$} is to use the \textit{regional similarities} (based on {\small$\bm{h}$}) as the hidden mechanism of mimicking the \textit{sample-wise similarity}.
Let {\small$x_1,x_2\in X$} be two samples, and the probability of {\small$x_2$} conditioned on {\small$x_1$} represents the sample-wise similarity of {\small$x_2$} to {\small$x_1$}.
Then, the objective of learning {\small$\Lambda$} is to minimize the KL divergence between the conditional probability {\small$P(x_2|x_1)$} inferred by the DNN and the conditional probability {\small$Q_{\Lambda}(x_2|x_1)$} inferred by the projected regional features {\small$\bm{h}$}.
In this way, each regional feature {\small$h^{(r)}$} can well reflect feature representation {\small$f^{(r)}$} used by the DNN.\vspace{-4pt}
\begin{small}
\begin{equation}
    \mathcal{L}_{\text{similarity}}\!=\!KL[P(X_2|X_1)\Vert Q_{\Lambda}(X_2|X_1)]
    \Rightarrow
    \frac{\partial\mathcal{L}_{\text{similarity}}}{\partial\Lambda}\!=\!-\mathbb{E}_{p_{\text{data}}(x_1)}\left[\mathbb{E}_{P(x_2|x=x_1)}\ \frac{\partial\log Q_{\Lambda}(x_2|x_1)}{\partial\Lambda}\right]
    \label{eq:r-emb-goal-func}
\end{equation}
\end{small}
{\small$P(x_2|x_1)$} reflects the similarity of {\small$x_2$} to {\small$x_1$} encoded by the DNN, which is computed using DNN's categorical outputs {\small$z_2,z_1\in\mathbb{R}^C$}.
We assume {\small$z_2$} follows a vMF distribution with mean direction {\small$z_1$}, \emph{i.e.} {\small$P(x_2|x_1)=\frac{1}{Z}\exp[\kappa_p\cdot\cos(z_2, z_1)]$}, where {\small$Z={\sum}_{x_2}\exp[\kappa_p\cdot\cos(z_2, z_1)]$}.

{\small$Q_{\Lambda}(x_2|x_1)$} reflects the similarity of {\small$x_2$} to {\small$x_1$} inferred by the projected regional features {\small$\bm{h}_2$} and {\small$\bm{h}_1$}.
Just like the bag-of-words model \cite{sivic2003video,csurka2004visual}, each projected regional feature {\small$h^{(r)}_2$} is assumed to independently contribute to the inference of {\small$Q_\Lambda(\bm{h}_2|\bm{h}_1)$} to simplify the computation.
Furthermore, {\small$h^{(r)}_2$} is weighted by its importance {\small$w^{(r)}_2>0$}, \emph{i.e.} {\small$Q_\Lambda(x_2|x_1)\propto\prod_r Q_\Lambda(h^{(r)}_2|\bm{h}_1)^{w^{(r)}_2}$}.
Just like in \cite{wang2017robust}, a large value of {\small$w^{(r)}_2$} means the {\small$r$}-th region in {\small$x_2$} is important for inference and peaks {\small$h^{(r)}_2$}'s contribution {\small$Q_\Lambda(h^{(r)}_2|\bm{h}_1)$}, while a weight {\small$w^{(r)}_2$} near zero flattens out {\small$Q_\Lambda(h^{(r)}_2|\bm{h}_1)$}.
Details of the estimation of {\small$w^{(r)}_2$} will be introduced later.
In this way, we have\vspace{-4pt}
\begin{small}
\begin{equation}
    \frac{\partial\log Q_\Lambda(x_2|x_1)}{\partial\Lambda}={\sum}_r w^{(r)}_2\ \frac{\partial\log Q_\Lambda(h^{(r)}_2|\bm{h}_1)}{\partial\Lambda},
    \label{eq:r-emb-decompose}
\end{equation}
\end{small}
where {\small$Q_\Lambda(h^{(r)}_2|\bm{h}_1)$} represents the likelihood of the sample {\small$x_1$} containing a regional feature {\small$h^{(r')}_1\in \bm{h}_1$}, that is similar to the regional feature {\small$h^{(r)}_2$} in sample {\small$x_2$}.
Thus, we compute {\small$Q_\Lambda(h^{(r)}_2|\bm{h}_1)$} as follows.\vspace{-4pt}
\begin{small}
\begin{equation}
    Q_\Lambda(h^{(r)}_2|\bm{h}_1)
    \!=\!Q_\Lambda(h^{(r)}_2|h^{(r')}_1)
    \!=\!p_{\text{vMF}}\left(\!h^{(r)}_2\Big|\mu\!=\!h^{(r')}_1\!,\kappa(\Vert h^{(r)}_2\Vert)\!\right)\!,
    \text{ s.t. }
    r'\!\!=\!\arg\max_{r'}Q_\Lambda(h^{(r)}_2|h^{(r')}_1)\!
    \label{eq:r-emb-single-region}
\end{equation}
\end{small}
Here, we assume {\small$h^{(r)}_2$} follows a revised vMF distribution in Eq. (\ref{eq:extended-vmf}) with mean direction {\small$h^{(r')}_1$}, where the {\small$r'$}-th region in {\small$x_1$} is selected as the most similar region to the {\small$r$}-th region in {\small$x_2$}.

As is shown above, the loss {\small$\mathcal{L}_{\text{similarity}}$} enables {\small$h^{(r)}$} to mimic feature representation of {\small$f^{(r)}$} in terms of encoding the sample-wise similarity.
Furthermore, we also expect {\small$h^{(r)}$} to reflect the discrimination power of each regional feature.
Therefore, we align the regional features {\small$\bm{h}$} to the coordinate system of {\small$g$} representing the sample-wise discrimination power, in order to represent the regional discrimination power.
To this end, we maximize the mutual information between the regional features and the sample features, as the second loss.
In other words, this loss enables the discrimination power of regional feature to infer that of the corresponding sample feature.\vspace{-4pt}
\begin{small}
\begin{equation}
    \mathcal{L}_{\text{align}}
    =-MI(\bm{h}(X);g(X))
    \Rightarrow
    \frac{\partial\mathcal{L}_{\text{align}}}{\partial\Lambda}=-\mathbb{E}_{Q_\Lambda(\bm{h},g)}\left[\frac{\partial\log Q_\Lambda(\bm{h}|g)}{\partial\Lambda}\right]
    \label{eq:r-emb-loss-2}
\end{equation}
\end{small}
The joint probability {\small$Q_\Lambda(\bm{h},g)=p(g)\cdot \prod_r Q_\Lambda(h^{(r)}|g)^{w^{(r)}}$} reflects the fitness between the discrimination power of a sample and that of its compositional regions.
{\small$Q_\Lambda(h^{(r)}|g)$} reflects the fitness between the sample feature {\small$g$} and each {\small$r$}-th regional feature {\small$h^{(r)}$}, which is assumed to follow a vMF distribution with mean direction {\small$g$}, \emph{i.e.} {\small$Q_\Lambda(h^{(r)}|g)=p_{\text{vMF}}(h^{(r)}|g,\kappa')$}.
In this way, the second loss can be equivalently written as {\small$\mathcal{L}_{\text{align}}=-\mathbb{E}_x[\sum_r w^{(r)}\cdot\cos(g,h^{(r)})]$}, where {\small$\kappa'$} has been eliminated (proof in the supplementary material).

In sum, the loss functions in Eq. (\ref{eq:r-emb-goal-func}) and (\ref{eq:r-emb-loss-2}) enable {\small$h^{(r)}$} to reflect both the feature representation of {\small$f^{(r)}$} and align {\small$h^{(r)}$} to the coordinate system of {\small$g$}'s discrimination power.
Thus, we learn {\small$\Lambda$} using both losses {\small$\mathcal{L}=\mathcal{L}_{\text{similarity}}+\alpha\cdot\mathcal{L}_{\text{align}}$ $(\alpha>0)$}.

\textbf{Estimation of each region's importance {\small$w^{(r)}$}}.
In Eq. (\ref{eq:r-emb-decompose}), we need to estimate the importance of each {\small$r$}-th region as {\small$w^{(r)}$}.
Just like Eq. (\ref{eq:r-emb-goal-func}), the objective of estimating {\small$\bm{w}=[w^{(1)},...,w^{(HW)}]$} in each sample is also formulated as the minimization of the KL divergence between {\small$P(x_2|x_1)$} inferred by the DNN and {\small$Q_{\bm{w}}(x_2|x_1)$} inferred by {\small$f$}.\vspace{-4pt}
\begin{small}
\begin{equation}
    \min_{\bm{w}}\ KL(P(X_2|X_1)\Vert Q_{\bm{w}}(X_2\Vert X_1))
    \Rightarrow
    \min_{\bm{w}}\ \mathbb{E}_{x_1}\left[{\sum}_{x_2}P(x_2|x_1)\log\frac{P(x_2|x_1)}{Q_{\bm{w}}(x_2|x_1)}\right]
    \label{eq:w-goal-func}
\end{equation}
\end{small}
Unlike Eq. (\ref{eq:r-emb-decompose}), we estimate {\small$w$} by formulating {\small$Q_{\bm{w}}(x_2|x_1)$} using raw features {\small$f$}, instead of the projected features {\small$\bm{h}$}, for more accurate estimation.
We assume each regional feature {\small$f^{(r)}_2$} contributes independently to {\small$Q_{\bm{w}}(x_2|x_1)$}, \emph{i.e.} {\small$ Q_{\bm{w}}(x_2|x_1)\propto{\prod}_r\  Q_{\bm{w}}(f^{(r)}_2|f_1)^{w^{(r)}_2}$}.
Then, just like Eq. (\ref{eq:r-emb-single-region}), {\small$Q_{\bm{w}}(f^{(r)}_2|f_1)=\max_{r'}Q_{\bm{w}}(f^{(r)}_2|f^{(r')}_1)$}.
In the quantification of {\small$Q_{\bm{w}}(f^{(r)}_2|f^{(r')}_1)$}, we further consider the different importance of each channel in {\small$f^{(r)}_2$}.
To this end,
we further estimate the importance of each channel of {\small$f_2$} as {\small$\bm{v}_2=[v_2^{(1)},...,v_2^{(K)}]\in\mathbb{R}^{K}$}, where {\small$v_2^{(k)}\in\mathbb{R}$} denotes the importance of the {\small$k$}-th channel.
In this way, we quantify {\small$Q_{\bm{w}}(f^{(r)}_2|f^{(r')}_1)$} as follows.
\begin{small}
\begin{equation}
    Q_{\bm{w}}(f^{(r)}_2|f^{(r')}_1)\propto \exp\left[\kappa'\cdot\sum_k\left(v^{(k)}_2\cdot\frac{f^{(r)}_{2,k}}{\Vert f^{(r)}_2\Vert_2}\cdot\frac{f^{(r)}_{1,k}}{\Vert f^{(r)}_1\Vert_2}\right)\right],
\end{equation}
\end{small}
where {\small$f^{(r)}_{2,k}$} and {\small$f^{(r)}_{1,k}$} are neural activations of the {\small$k$}-th channel in the {\small$r$}-th region in {\small$f_2$} and {\small$f_1$}, respectively.
In our experiments, {\small$\bm{w}_2$} and {\small$\bm{v}_2$} were jointly optimized via Eq. (\ref{eq:w-goal-func}).
For fair comparison between different samples, we force each element in {\small$\bm{w}_2$} and {\small$\bm{v}_2$} to be non-negative, and force their {\small$L_1$}-norm to be 1. \emph{I.e.} {\small$\bm{w}_2\succeq 0$}, {\small$\bm{v}_2\succeq 0$}, {\small$\Vert\bm{w}_2\Vert_1=1$}, and {\small$\Vert\bm{v}_2\Vert_1=1$}.
This ensures that magnitudes of region's/channel's importance in different samples are similar. Besides, this constraint can also stabilize the optimization process of {\small$\bm{w}_2$} and {\small$\bm{v}_2$}.
The optimization process of {\small$\bm{w}_2$} and {\small$\bm{v}_2$} alternates between the following two steps. (i) We first update {\small$\bm{w}_2$} and {\small$\bm{v}_2$} via Eq. (\ref{eq:w-goal-func}) using the gradient descent method. (ii) We force each element in {\small$\bm{w}_2$} and {\small$\bm{v}_2$} to be non-negative and force their {\small$L_1$}-norm to be 1, \emph{i.e.}
the importance of the {\small$r$}-th region {\small$w^{(r)}_2$} is normalized to
{\small$\frac{|w^{(r)}_2|}{\Vert\bm{w}_2\Vert_1}$} {\small$(r=1,...,HW)$}, and the importance of the {\small$k$}-th channel {\small$v^{(k)}_2$} is normalized to {\small$\frac{|v^{(k)}_2|}{\Vert\bm{v}_2\Vert_1}$} {\small$(k=1,...,K)$}.

\begin{figure}[t]
\begin{minipage}[t]{0.38\linewidth}
\vspace{0pt}
    \centering
    \includegraphics[width=.95\linewidth]{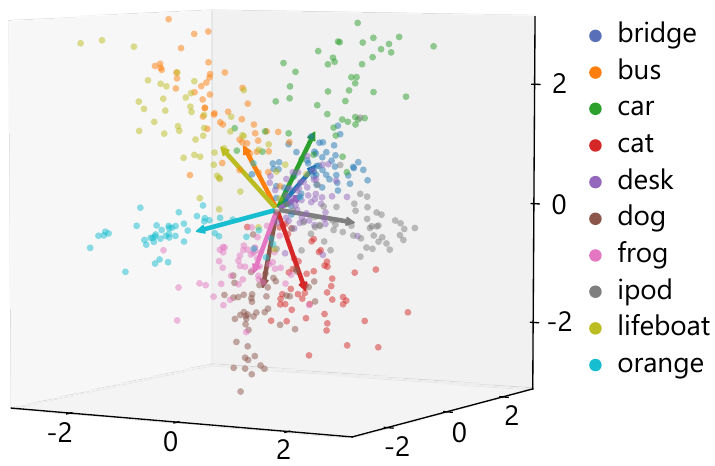}
    \vspace{-2pt}
    \caption{\small Visualization of sample features learned by VGG-16 in the coordinate system of the Tiny ImageNet categories\textsuperscript{1}.}
    \label{fig:sample-emb-final}
\end{minipage}
\hfill
\begin{minipage}[t]{0.6\linewidth}
\vspace{5pt}
    \centering
    \renewcommand{\arraystretch}{1.5}
    \begin{small}
    \resizebox{\textwidth}{!}{
    \begin{tabular}{c|ccc|c|c}
    \hline
    dataset & \multicolumn{3}{c|}{Tiny ImageNet}  & COCO 2014 & CUB-200-2011 \\
    DNN  & VGG-16 & ResNet-34 & MobileNet-V2 & ResNet-50 & ResNet-34    \\
    \hline
    PCA & -0.65 & -0.78 & -0.81 & -0.56 & -0.78 \\
    t-SNE & -0.50 & -0.36 & -0.66 & -0.67 & -0.50 \\
    LLE & -0.38 & -0.51 & -0.27 & -0.09 & -0.58 \\
    ISOMAP & -0.66 & -0.83 & -0.77 & -0.65 & -0.75 \\
    DRPR &  -0.87 & -0.89 & -0.88 & -0.77 & -0.77 \\
    \hline
    ours & \textbf{-0.94} & \textbf{-0.94} & \textbf{-0.95} & \textbf{-0.84} & \textbf{-0.90} \\
    \hline
    \end{tabular}%
    }
    \end{small}
    \renewcommand{\arraystretch}{1}
    \captionof{table}{{\small The negative correlation ({\scriptsize$\downarrow$}) between the visualized sample features' strength and the samples' classification uncertainty.}}
    \label{tab:rel-s-disc-strength}
\end{minipage}
\vspace{-1.2em}
\end{figure}


\subsection{Quantifying knowledge points and the ratio of reliable knowledge points}

Visualizing the discrimination power of regional features provides us a new perspective to analyze the representation capacity of a DNN, \emph{i.e.} counting knowledge points encoded in different layers, and quantifying the ratio of reliable knowledge points.
Up to now, \citet{cheng2020explaining} was the first attempt to quantify the knowledge points encoded in an intermediate layer using the information theory, but the knowledge points were extracted based on the discard of the pixel-wise information, instead of representing the discrimination power of regional features.
In comparison, we quantify the total knowledge points and reliable ones, in terms of their discrimination power.
Experiments show that the quantity and quality of knowledge points can well explain knowledge distillation in practice.

Specifically, given a regional feature {\small$h^{(r)}$}, if {\small$h^{(r)}$} is discriminative enough for classification of any category, \emph{i.e.} {\small$\max_c p(y=c|h^{(r)})>\tau$}, then we count this regional feature as a \textit{knowledge point}.
The classification probability {\small$p(y=c|h^{(r)})=\frac{\pi_c\cdot\exp[\kappa(\Vert h^{(r)}\Vert_2)\cdot\cos(h^{(r)}, \mu_c)]}{\sum_{c'}\pi_{c'}\cdot\exp[\kappa(\Vert h^{(r)}\Vert_2)\cdot\cos(h^{(r)}, \mu_{c'})]}$}, which is similar to Eq. (\ref{eq:posterior}).
Furthermore, among all knowledge points, those pushing the classification towards the correct classification, \emph{i.e.} knowledge points satisfying {\small$c^{\text{truth}}=\arg\max_c p(y=c|h^{(r)})$}, are taken as \textit{reliable knowledge points.}
In this way, \textit{the ratio of reliable knowledge points} is defined as the ratio of reliable knowledge points to total knowledge points, which reflects the quality of visual patterns.

\begin{figure}[t]
    \centering
    \includegraphics[width=\linewidth]{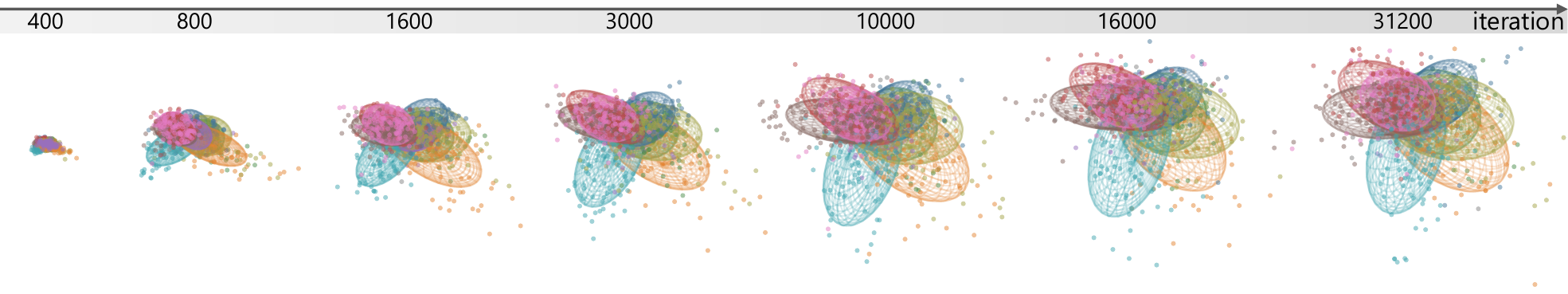}
    \vspace{-23pt}
    \caption{{\small The emergence of regional patterns through the learning process in the coordinate system of visualizing the Tiny ImageNet categories\textsuperscript{\ref{fn:dataset}}.}}
    \vspace{-12pt}
    \label{fig:r-emergence-training}
\end{figure}

\begin{figure}[t]
    \centering
    \includegraphics[width=\linewidth]{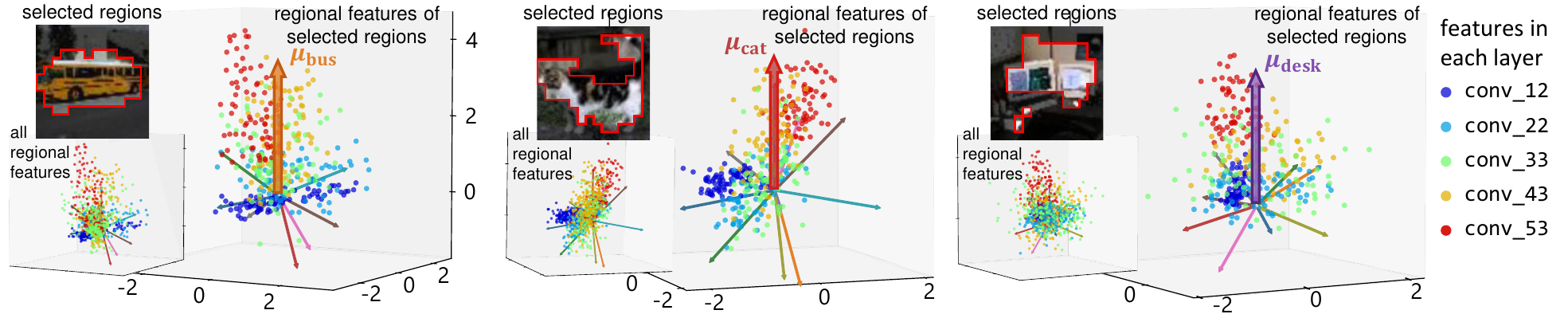}
    \vspace{-13pt}
    \caption{\small The emergence of regional patterns through the forward propagation in the coordinate system of visualizing the Tiny ImageNet categories\textsuperscript{\ref{fn:dataset}}. Coordinates along the vertical axis reflect the discrimination power of the target category.}
    \vspace{-16pt}
    \label{fig:r-emergence-layer}
\end{figure}

\section{Experiments}
\label{sec:exp}

In this section, we used our method to visualize sample features and regional features in VGG-16~\cite{simonyan2014very}, ResNet-34/50~\cite{he2016deep}, MobileNet-V2~\cite{sandler2018mobilenetv2}, which were learned for object classification, based on the Tiny ImageNet dataset~\cite{le2015tiny}, the MS COCO 2014 dataset~\cite{lin2014microsoft}, and the CUB-200-2011 dataset~\cite{WahCUB_200_2011}.
For the MS COCO 2014 dataset and the CUB-200-2011 dataset, we used images cropped by the annotated bounding boxes for both training and testing.
Note that the analysis of classification for massive categories requires a large number of category directions in the coordinate system, which will hurt the visualization clarity of the radial distribution.
Therefore, to clarify the visualization result, we randomly selected 10 categories from each dataset\footnote{For the Tiny ImageNet dataset, we selected \textit{steel arch bridge, school bus, sports car, tabby cat, desk, golden retriever, tailed frog, iPod, lifeboat, and orange}. Please see the supplementary material for other datasets.\label{fn:dataset}}.
Please see the supplementary material for details on the DNNs and datasets.

\textbf{Visualization and verification of sample features' discrimination power.}
In this experiment, we projected sample features {\small$f$} into a 3-dimensional space (\emph{i.e.} {\small$d'\!=\!3$}) for visualization.
Specifically, we selected the feature before the last fully-connected layer as the sample feature {\small$f$}.
Figure \ref{fig:sample-emb-final} shows the projected sample features {\small$g$} and each category direction {\small$\mu_c$}.
The visualization result revealed the semantic similarity between categories.
For example, \textit{cat} features were similar to \textit{dog} features, and \textit{bus} features were similar to \textit{lifeboat} features.
Besides, the supplementary material shows how the discrimination power of sample features gradually increased through the training process.

Furthermore, in order to examine whether {\small$g$} reflected the discrimination power of sample features, we evaluated the Pearson correlation coefficient between the strength {\small$\Vert g\Vert_2$} and the classification uncertainty of each sample {\small$x$}.
To this end, the classification uncertainty of each sample {\small$x$} was measured as the entropy of its output probability, \emph{i.e.} {\small$H(Y|X=x)$}.
In Table \ref{tab:rel-s-disc-strength}, we compared our method with several visualization methods, such as PCA~\cite{pearson1901liii}, t-SNE~\cite{van2008visualizing}, LLE~\cite{roweis2000nonlinear}, ISOMAP~\cite{tenenbaum2000global}, and the recent DRPR~\cite{law2019dimensionality}.
\textit{Compared with baseline methods, the strength of our projected sample features {\small$g$} was more strongly correlated to the classification uncertainty.}
The supplementary material also verified the effectiveness of our method by illustrating contour maps of the classification probability.

\textbf{Emergence of discriminative regional features.}
Next, we projected regional features {\small$f$} into a 3-dimension space (\emph{i.e.} {\small$d'\!=\!3$}) to analyze the importance and reliability of each {\small$f^{(r)}$} towards classification.
We set {\small$\alpha=0.1$}.
Figure \ref{fig:r-emergence-training} shows the emergence of projected regional features {\small$h^{(r)}$} through the training process, when we selected the output feature of the {\small\texttt{conv\_53}} layer of VGG-16 as {\small$f^{(r)}$}.
The ellipsoid represented the estimated Gaussian distribution of {\small$h^{(r)}$} for image regions cropped from each category.
The visualization result showed that the discrimination power and reliability of regional features gradually increased during training.
Besides, Figure \ref{fig:r-emergence-layer} visualizes regional features extracted from different layers of VGG-16.
For clarity, we further selected image regions corresponding to reliable knowledge points in the {\small\texttt{conv\_53}} layer.
Figure \ref{fig:r-emergence-layer} visualizes the selected regions, as well as their regional features.
It showed that these regions were not discriminative in low layers, but became discriminative in high layers.

\begin{figure}[t]
\centering
\begin{minipage}[c]{.63\linewidth}
\centering
    \includegraphics[width=.95\linewidth]{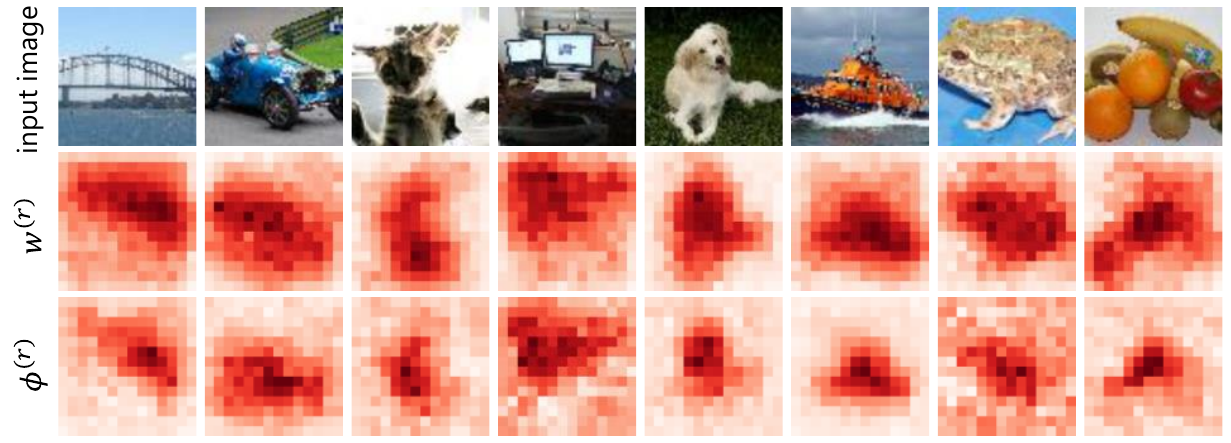}
    \vspace{-6pt}
    \caption{\small Visualization of the regional importance estimated by our method.
    Our regional importance is similar to the Shapley value of {\footnotesize$f^{(r)}$}, which verifies the trustworthiness of our method.}
    \label{fig:importance-shap}
\end{minipage}
\hfill
\begin{minipage}[c]{.36\linewidth}
    \renewcommand{\arraystretch}{1.4}
    \resizebox{\textwidth}{!}{
    \begin{small}
    \begin{tabular}{c|c|c}
    \hline
    dataset    & DNN & correlation \\ \hline
    \multirow{3}{*}{Tiny ImageNet} & VGG-16 & 0.7707{\scriptsize$\pm$0.16} \\
                  & ResNet-34 & 0.8248{\scriptsize$\pm$0.09} \\
                  & MobileNet-V2 & 0.8169{\scriptsize$\pm$0.13} \\ \hline
    COCO 2014   & ResNet-50 & 0.7572{\scriptsize$\pm$0.18} \\ \hline
    CUB-200-2011  & ResNet-34 & 0.7765{\scriptsize$\pm$0.17} \\ \hline
\end{tabular}%
\end{small}
}
\renewcommand{\arraystretch}{1}
\vspace{-1pt}
\captionof{table}{{\small The Pearson correlation coefficient between {\small$\Vert h^{(r)}\Vert_2$} and {\small$w^{(r)}$}. The feature strength and feature importance were positively related to each other.}}
\label{tab:importance-strength}
\end{minipage}
\vspace{-1em}
\end{figure}

\textbf{Visualization and verification of the estimated regional importance.}
In this experiment, we estimated regional importance {\small$w^{(r)}$} with {\small$\tilde{\kappa}$} set to 1000.
The estimated {\small$w^{(r)}$} was further verified from the following two perspectives.
From the first perspective, we compared the estimated regional importance {\small$w^{(r)}$} and the Shapley value~\cite{shapley1953value,lunberg2017unified} {\small$\phi^{(r)}$} of each {\small$r$}-th region, when we selected the output feature of the {\small\texttt{conv\_53}} layer of VGG-16 as regional features {\small$f^{(r)}$}.
To this end, the Shapley value {\small$\phi^{(r)}$} was computed as the numerical contribution of {\small$f^{(r)}$} to the DNN output.
The Shapley value is the unique unbiased and widely-used~\cite{chen2018lshapley,amirata2019data,williamson2020efficient} metric that fairly allocates the numerical contribution to input features, which satisfies the linearity axiom, the dummy axiom, the symmetry axiom, and the efficiency axiom~\cite{ancona2019explaining}.
Figure \ref{fig:importance-shap} shows the high similarity between {\small$w^{(r)}$} and {\small$\phi^{(r)}$} among different regions {\small$r$}, which demonstrated the trustworthiness of the estimated regional importance {\small$w^{(r)}$}.

Besides, we calculated the Pearson correlation coefficient between the strength of projected features {\small$\Vert h^{(r)}\Vert_2$} and their corresponding importance {\small$w^{(r)}$}.
Table~\ref{tab:importance-strength} shows the mean value and the standard deviation of the Pearson correlation coefficient through all input samples in each dataset, when we used the output feature of the last convolutional layer as regional features {\small$f$}.
This proved that feature strength and feature importance were significantly and positively related to each other.

\begin{figure}[t]
    \centering
    \includegraphics[width=\linewidth]{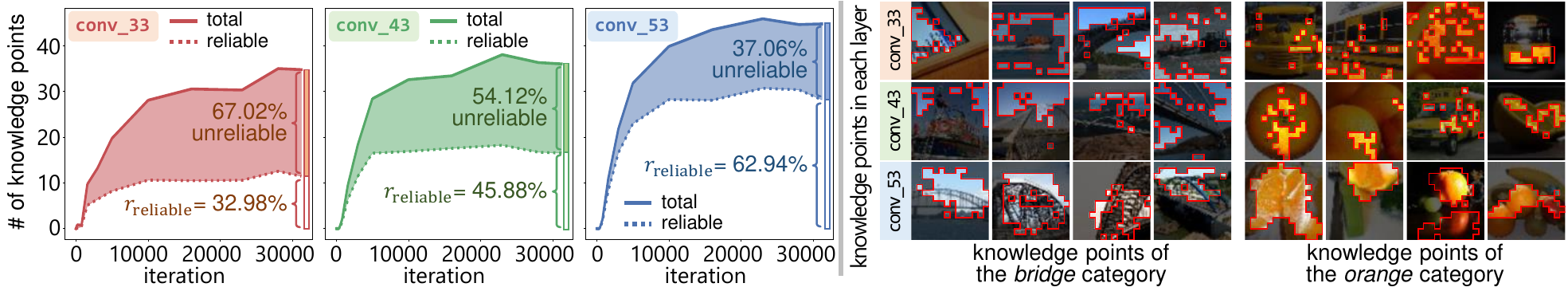}
    \vspace{-15pt}
    \caption{\small (left) The increase of total knowledge points and reliable knowledge points during training. The ratio of reliable knowledge points, {\small$r_{\text{reliable}}\!=\!{\text{\# of reliable points}/\text{\# of all points}}$}, increases through the forward propagation. (right) Visualization of image regions of knowledge points towards different categories.}
    \vspace{-1.5em}
    \label{fig:quantify-concept}
\end{figure}

\textbf{Quantifying knowledge points and the ratio of reliable knowledge points.}
Figure \ref{fig:quantify-concept}(left) shows the increase of knowledge points in different layers through the training of VGG-16.
For fair comparison, we normalized the average strength of regional features {\small$\mathbb{E}_{x,r}[\Vert h^{(r)}\Vert_2\ _{\text{given }x}]$} in each layer to the average strength of regional features in the {\small\texttt{conv\_53}} layer, and therefore we could simply set {\small$\tau=0.4$}.
Besides, we also computed the ratio of reliable knowledge points in each layer.
Figure \ref{fig:quantify-concept}(left) shows that the ratio of reliable knowledge points in high layers (\emph{e.g.} the {\small\texttt{conv\_53}} layer) was higher than that in low layers (\emph{e.g.} the {\small\texttt{conv\_33}} layer), which demonstrated the increasing quality of visual patterns through the forward propagation.
Besides, Figure \ref{fig:quantify-concept}(right) highlights the image regions of knowledge points towards different categories.
Regional features in high layers were usually more likely to be localized on the foreground than regional features in low layers.

\begin{figure}[t]
    \centering
    \includegraphics[width=\linewidth]{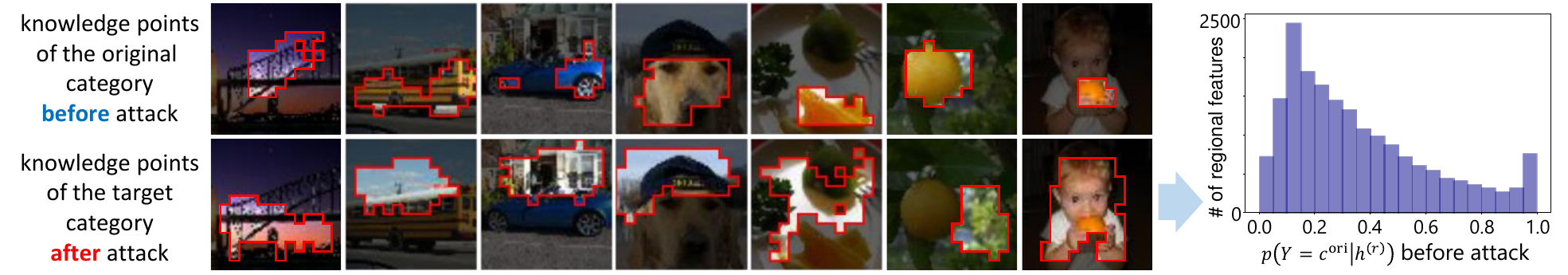}
    \vspace{-1em}
    \caption{{\small (left) Image regions corresponding to knowledge points in original and adversarial samples.
    (right) We selected important regions for the target category from adversarial samples, and evaluated the selected regions' utilities/importance of classifying original images to the true category. Most important regions after the attack were not so important before the attack.}}
    \vspace{-1em}
    \label{fig:adv-region-vis-hist}
\end{figure}

\begin{figure}[t]
    \centering
    \includegraphics[width=\linewidth]{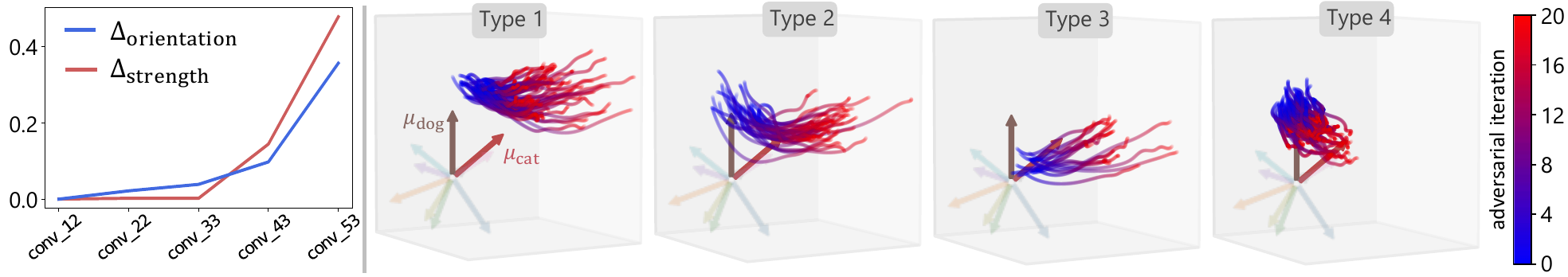}
    \vspace{-1.3em}
    \caption{{\small (left) The adversarial attack usually made significant effects on regional features in high layers. (right) Visualization of four types of regional features' trajectories during the attack.}}
    \vspace{-1.5em}
    \label{fig:layer-trajectory}
\end{figure}

\begin{figure}[t!]
\centering
\begin{minipage}[c]{.59\linewidth}
    \includegraphics[width=\linewidth]{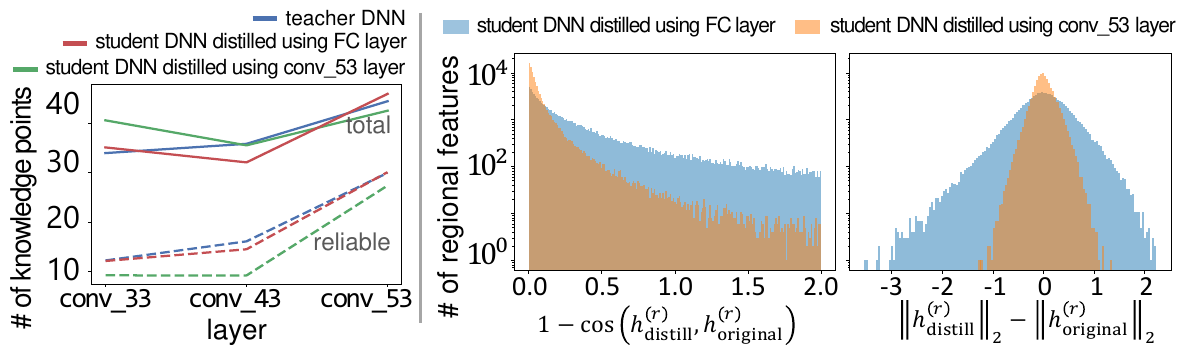}
\end{minipage}
\hfill
\begin{minipage}[c]{.4\linewidth}
\vspace{10pt}
    \caption{{\small(left) Knowledge distillation caused the DNN to encode less reliable knowledge points. (right) The dissimilarity of features between each student DNN and the teacher DNN, in terms of orientation and strength.}}
    \label{fig:distillation-compare}
\end{minipage}
\vspace{-1.6em}
\end{figure}

\textbf{The adversarial attack mainly affected unreliable regional features in high layers.}
We used our method to analyze the change of regional features when we applied the adversarial attack~\cite{madry2017towards} to VGG-16.
Given a normal sample {\small$x$}, the adversarial sample {\small$x_{\text{adv}}=x+\delta$} was generated via the untargeted  PGD attack~\cite{madry2017towards}, subject to {\small$\Vert\delta\Vert_\infty\leq \frac{1}{255}$}. The attack was iterated for {\small$20$} steps with the step size of {\small$\frac{0.1}{255}$}.
In Figure \ref{fig:adv-region-vis-hist}(left), we found that important regions for the classification of the original image (the first row) were usually different from important regions that attacked the classification towards the target category (the second row).
More specifically, let {\small$h^{(r)}_{\text{ori}}$} and {\small$h^{(r)}_{\text{adv}}$} denote two corresponding regional features in the same layer before and after the attack.
Let us select important regions {\small$\{r\}$} for attacking from adversarial samples, satisfying {\small$p(y=c^{\text{adv}}|h^{(r)}_{\text{adv}})>0.4$}.
Figure~\ref{fig:adv-region-vis-hist}(right) illustrates the histogram for the selected regions' classification utilities {\small$p(y=c^{\text{ori}}|h^{(r)}_{\text{ori}})$} in the original image.
We found that most important regions after the attack were not so important before the attack.
Besides, we compared the utility of the attack to regional features in different layers.
Let {\small$\Delta_{\text{orientation}}=\mathbb{E}_{x}[\mathbb{E}_r(\cos(h^{(r)}_{\text{ori}},h^{(r)}_{\text{adv}}))]$} and {\small$\Delta_{\text{strength}}=\mathbb{E}_{x}[\mathbb{E}_r(|\Vert h^{(r)}_{\text{ori}}\Vert_2-\Vert h^{(r)}_{\text{adv}}\Vert_2|)]$} measure the utility of the attack to regional features' orientation and strength.
Figure~\ref{fig:layer-trajectory} shows that the adversarial attack mainly affected regional features in high layers, \emph{e.g.} the {\small\texttt{conv\_53}} layer in VGG-16.
We further categorized all image regions into four types, in terms of their attacking behaviors.
To this end, we visualized the trajectories of regional features in the {\small\texttt{conv\_53}} during the attack.
As Figure~\ref{fig:layer-trajectory}(right) shows, Type 1 illustrates important image regions for the \textit{dog} category that were directly transferred to the \textit{cat} category without much difficulties.
Type 2 illustrates important \textit{dog} regions, in which \textit{dog} features were first damaged and then \textit{cat} features were built up and became important \textit{cat} regions. Type 3 indicates unimportant \textit{dog} regions that were pushed to important \textit{cat} regions. Type 4 indicates important \textit{dog} regions that were damaged by the attack and became unimportant regions.

\textbf{The DNN learned via knowledge distillation encoded less reliable visual patterns.}
In this experiment, we learned two \textit{student DNN}s (two VGG-16 nets) for knowledge distillation~\cite{hinton2015distilling}.
One \textit{student DNN} was learned by distilling the output feature of the {\small\texttt{conv\_53}} layer after the ReLU operation in the \textit{teacher DNN} (a pre-trained VGG-16) to the corresponding layer in the student DNN.
The other \textit{student DNN} was learned by distilling the output feature of the penultimate fully-connected layer after the ReLU operation in the teacher DNN to the corresponding layer in the student DNN.
Figure \ref{fig:distillation-compare}(left) compares the number of all knowledge points and reliable knowledge points encoded by the teacher DNN and the two student DNNs, when we quantified knowledge points in {\small\texttt{conv\_33/conv\_43/conv\_53}} layers.
We found that \textit{student DNNs usually encoded less reliable knowledge points than the teacher DNN.}

$\bullet$ Furthermore, \textit{the student DNN usually learned even less reliable concepts in a layer, if this layer was farther from the target layer used for distillation.}
To verify this conclusion, we compared the number of knowledge points between the above two student DNNs.
As Figure \ref{fig:distillation-compare}(left) shows, the student DNN distilled using features of the fully-connected layer encoded much less reliable concepts than the student DNN distilled using features of the {\small\texttt{conv\_53}} layer, which verified our conclusion.

$\bullet$ \textit{Although the knowledge distillation could force the student DNN to well mimic features of a specific layer in teacher DNN, there was still a big difference of other layers' regional features between the student DNN and the teacher DNN.}
To this end, we evaluated the quality of student DNNs mimicking the teacher DNN.
We selected {\small$h^{(r)}_{\text{student}}$} and {\small$h^{(r)}_{\text{teacher}}$} as two corresponding regional features of the student DNN and the teacher DNN in the same layer.
Then, we used {\small$1-\cos(h^{(r)}_{\text{student}},h^{(r)}_{\text{teacher}})$} and {\small$\Vert h^{(r)}_{\text{student}}\Vert_2-\Vert h^{(r)}_{\text{teacher}}\Vert_2$} to measure the difference of orientation and the difference of strength between the two regional features.
Figure \ref{fig:distillation-compare}(right) shows the histogram of {\small$1-\cos(h^{(r)}_{\text{student}},h^{(r)}_{\text{teacher}})$} and {\small$\Vert h^{(r)}_{\text{student}}\Vert_2-\Vert h^{(r)}_{\text{teacher}}\Vert_2$}, when we used the {\small\texttt{conv\_53}} layer to evaluate the similarity between the student DNN and the teacher DNN.
The similarity between student DNN features and teacher DNN features was lower when the student DNN was distilled based on features in the fully-connected layer (far from the {\small\texttt{conv\_53}} layer), which verified our conclusion.



\section{Conclusion}

In this paper, we propose a method to visualize intermediate visual patterns in a DNN.
The visualization illustrates the emergence of intermediate visual patterns in a temporal-spatial manner.
The proposed method also enables people to measure the quantity and quality of visual patterns encoded by the DNN, which provides a new perspective to analyze the discrimination power of DNNs.
Furthermore, the proposed method provides insightful understanding towards the signal-processing behaviors of existing deep-learning techniques.

\begin{ack}
This work is partially supported by the National Nature Science Foundation of China (No. 61906120, U19B2043), Shanghai Natural Science Fundation (21JC1403800,21ZR1434600), Shanghai Municipal Science and Technology Major Project (2021SHZDZX0102).
\end{ack}

\bibliographystyle{plainnat}
\bibliography{ref}

\begin{thebibliography}{82}
\providecommand{\natexlab}[1]{#1}
\providecommand{\url}[1]{\texttt{#1}}
\expandafter\ifx\csname urlstyle\endcsname\relax
  \providecommand{\doi}[1]{doi: #1}\else
  \providecommand{\doi}{doi: \begingroup \urlstyle{rm}\Url}\fi

\bibitem[Abramowitz et~al.(1972)Abramowitz, Stegun,
  et~al.]{abramowitz1972handbook}
Milton Abramowitz, Irene~A Stegun, et~al.
\newblock \emph{Handbook of mathematical functions: with formulas, graphs, and
  mathematical tables}, volume~55.
\newblock National bureau of standards Washington, DC, 1972.

\bibitem[Achille and Soatto(2018)]{achille2018information}
Alessandro Achille and Stefano Soatto.
\newblock Information dropout: Learning optimal representations through noisy
  computation.
\newblock \emph{IEEE transactions on pattern analysis and machine
  intelligence}, 40\penalty0 (12):\penalty0 2897--2905, 2018.

\bibitem[Ancona et~al.(2019)Ancona, Oztireli, and Gross]{ancona2019explaining}
Marco Ancona, Cengiz Oztireli, and Markus Gross.
\newblock Explaining deep neural networks with a polynomial time algorithm for
  shapley values approximation.
\newblock \emph{arXiv:1903.10992}, 2019.

\bibitem[Banerjee et~al.(2005{\natexlab{a}})Banerjee, Dhillon, Ghosh, Sra, and
  Ridgeway]{banerjee2005clusteringvmf}
Arindam Banerjee, Inderjit~S Dhillon, Joydeep Ghosh, Suvrit Sra, and Greg
  Ridgeway.
\newblock Clustering on the unit hypersphere using von mises-fisher
  distributions.
\newblock \emph{Journal of Machine Learning Research}, 6\penalty0 (9),
  2005{\natexlab{a}}.

\bibitem[Banerjee et~al.(2005{\natexlab{b}})Banerjee, Merugu, Dhillon, Ghosh,
  and Lafferty]{banerjee2005clusteringbregman}
Arindam Banerjee, Srujana Merugu, Inderjit~S Dhillon, Joydeep Ghosh, and John
  Lafferty.
\newblock Clustering with bregman divergences.
\newblock \emph{Journal of machine learning research}, 6\penalty0 (10),
  2005{\natexlab{b}}.

\bibitem[Chatterji et~al.(2019)Chatterji, Neyshabur, and
  Sedghi]{chatterji2019intriguing}
Niladri~S Chatterji, Behnam Neyshabur, and Hanie Sedghi.
\newblock The intriguing role of module criticality in the generalization of
  deep networks.
\newblock \emph{arXiv preprint arXiv:1912.00528}, 2019.

\bibitem[Chattopadhay et~al.(2018)Chattopadhay, Sarkar, Howlader, and
  Balasubramanian]{chattopadhay2018grad}
Aditya Chattopadhay, Anirban Sarkar, Prantik Howlader, and Vineeth~N
  Balasubramanian.
\newblock Grad-cam++: Generalized gradient-based visual explanations for deep
  convolutional networks.
\newblock In \emph{2018 IEEE Winter Conference on Applications of Computer
  Vision (WACV)}, pages 839--847. IEEE, 2018.

\bibitem[Chen et~al.(2019)Chen, Song, Wainwright, and Jordan]{chen2018lshapley}
Jianbo Chen, Le~Song, Martin~J. Wainwright, and Michael~I. Jordan.
\newblock L-shapley and c-shapley: Efficient model interpretation for
  structured data.
\newblock \emph{In {ICLR}}, 2019.

\bibitem[Cheng et~al.(2020)Cheng, Rao, Chen, and Zhang]{cheng2020explaining}
Xu~Cheng, Zhefan Rao, Yilan Chen, and Quanshi Zhang.
\newblock Explaining knowledge distillation by quantifying the knowledge.
\newblock In \emph{Proceedings of the IEEE/CVF Conference on Computer Vision
  and Pattern Recognition (CVPR)}, June 2020.

\bibitem[Cheng et~al.(2021{\natexlab{a}})Cheng, Chu, Zheng, Ren, and
  Zhang]{cheng2021concepts}
Xu~Cheng, Chuntung Chu, Yi~Zheng, Jie Ren, and Quanshi Zhang.
\newblock A game-theoretic taxonomy of visual concepts in dnns.
\newblock \emph{arXiv preprint arXiv:2106.10938}, 2021{\natexlab{a}}.

\bibitem[Cheng et~al.(2021{\natexlab{b}})Cheng, Wang, Xue, Liang, and
  Zhang]{cheng2021aesthetic}
Xu~Cheng, Xin Wang, Haotian Xue, Zhengyang Liang, and Quanshi Zhang.
\newblock A hypothesis for the aesthetic appreciation in neural networks.
\newblock \emph{arXiv preprint arXiv:2108.02646}, 2021{\natexlab{b}}.

\bibitem[Cox and Cox(2008)]{cox2008multidimensional}
Michael~AA Cox and Trevor~F Cox.
\newblock Multidimensional scaling.
\newblock In \emph{Handbook of data visualization}, pages 315--347. Springer,
  2008.

\bibitem[Csurka et~al.(2004)Csurka, Dance, Fan, Willamowski, and
  Bray]{csurka2004visual}
Gabriella Csurka, Christopher Dance, Lixin Fan, Jutta Willamowski, and
  C{\'e}dric Bray.
\newblock Visual categorization with bags of keypoints.
\newblock In \emph{Workshop on statistical learning in computer vision, ECCV}.
  Prague, 2004.

\bibitem[Dierckx(1995)]{dierckx1995curve}
Paul Dierckx.
\newblock \emph{Curve and surface fitting with splines}.
\newblock Oxford University Press, 1995.

\bibitem[Dosovitskiy and Brox(2016)]{dosovitskiy2016inverting}
Alexey Dosovitskiy and Thomas Brox.
\newblock Inverting visual representations with convolutional networks.
\newblock In \emph{Proceedings of the IEEE conference on computer vision and
  pattern recognition}, pages 4829--4837, 2016.

\bibitem[Du et~al.(2018)Du, Wang, Zhai, Balakrishnan, Salakhutdinov, and
  Singh]{du2018many}
Simon~S Du, Yining Wang, Xiyu Zhai, Sivaraman Balakrishnan, Ruslan
  Salakhutdinov, and Aarti Singh.
\newblock How many samples are needed to estimate a convolutional neural
  network?
\newblock In \emph{NeurIPS}, pages 371--381, 2018.

\bibitem[Fisher et~al.(1993)Fisher, Lewis, and Embleton]{fisher1993statistical}
Nicholas~I Fisher, Toby Lewis, and Brian~JJ Embleton.
\newblock \emph{Statistical analysis of spherical data}.
\newblock Cambridge university press, 1993.

\bibitem[Fisher(1953)]{fisher1953dispersion}
Ronald~Aylmer Fisher.
\newblock Dispersion on a sphere.
\newblock \emph{Proceedings of the Royal Society of London. Series A.
  Mathematical and Physical Sciences}, 217\penalty0 (1130):\penalty0 295--305,
  1953.

\bibitem[Fong and Vedaldi(2017)]{fong2017interpretable}
Ruth~C Fong and Andrea Vedaldi.
\newblock Interpretable explanations of black boxes by meaningful perturbation.
\newblock In \emph{Proceedings of the IEEE International Conference on Computer
  Vision}, pages 3429--3437, 2017.

\bibitem[Fort et~al.(2019)Fort, Nowak, and Narayanan]{fort2019stiffness}
Stanislav Fort, Pawe{\l}~Krzysztof Nowak, and Srini Narayanan.
\newblock Stiffness: A new perspective on generalization in neural networks.
\newblock \emph{arXiv preprint arXiv:1901.09491}, 2019.

\bibitem[Ghorbani and Zou(2019)]{amirata2019data}
Amirata Ghorbani and James Zou.
\newblock Data shapley: Equitable valuation of data for machine learning.
\newblock \emph{In {ICML}}, 2019.

\bibitem[Goh et~al.(2021)Goh, Cammarata, Voss, Carter, Petrov, Schubert,
  Radford, and Olah]{goh2021multimodal}
Gabriel Goh, Nick Cammarata, Chelsea Voss, Shan Carter, Michael Petrov, Ludwig
  Schubert, Alec Radford, and Chris Olah.
\newblock Multimodal neurons in artificial neural networks.
\newblock \emph{Distill}, 2021.
\newblock \doi{10.23915/distill.00030}.
\newblock https://distill.pub/2021/multimodal-neurons.

\bibitem[Goldfeld et~al.(2019)Goldfeld, Van Den~Berg, Greenewald, Melnyk,
  Nguyen, Kingsbury, and Polyanskiy]{goldfeld2019estimating}
Ziv Goldfeld, Ewout Van Den~Berg, Kristjan Greenewald, Igor Melnyk, Nam Nguyen,
  Brian Kingsbury, and Yury Polyanskiy.
\newblock Estimating information flow in deep neural networks.
\newblock In \emph{International Conference on Machine Learning}, pages
  2299--2308, 2019.

\bibitem[Harley(2015)]{harley2015interactive}
Adam~W Harley.
\newblock An interactive node-link visualization of convolutional neural
  networks.
\newblock In \emph{International Symposium on Visual Computing}, pages
  867--877. Springer, 2015.

\bibitem[Hasnat et~al.(2017)Hasnat, Bohn{\'e}, Milgram, Gentric, Chen,
  et~al.]{hasnat2017mises}
Md~Hasnat, Julien Bohn{\'e}, Jonathan Milgram, St{\'e}phane Gentric, Liming
  Chen, et~al.
\newblock von mises-fisher mixture model-based deep learning: Application to
  face verification.
\newblock \emph{arXiv preprint arXiv:1706.04264}, 2017.

\bibitem[He et~al.(2016)He, Zhang, Ren, and Sun]{he2016deep}
Kaiming He, Xiangyu Zhang, Shaoqing Ren, and Jian Sun.
\newblock Deep residual learning for image recognition.
\newblock In \emph{Proceedings of the IEEE conference on computer vision and
  pattern recognition}, pages 770--778, 2016.

\bibitem[Hinton et~al.(2015)Hinton, Vinyals, and Dean]{hinton2015distilling}
Geoffrey Hinton, Oriol Vinyals, and Jeff Dean.
\newblock Distilling the knowledge in a neural network.
\newblock \emph{arXiv preprint arXiv:1503.02531}, 2015.

\bibitem[Kim et~al.(2018)Kim, Wattenberg, Gilmer, Cai, Wexler, Viegas,
  et~al.]{kim2018interpretability}
Been Kim, Martin Wattenberg, Justin Gilmer, Carrie Cai, James Wexler, Fernanda
  Viegas, et~al.
\newblock Interpretability beyond feature attribution: Quantitative testing
  with concept activation vectors (tcav).
\newblock In \emph{International conference on machine learning}, pages
  2668--2677. PMLR, 2018.

\bibitem[Kindermans et~al.(2017)Kindermans, Sch{\"u}tt, Alber, M{\"u}ller,
  Erhan, Kim, and D{\"a}hne]{kindermans2017learning}
Pieter-Jan Kindermans, Kristof~T Sch{\"u}tt, Maximilian Alber, Klaus-Robert
  M{\"u}ller, Dumitru Erhan, Been Kim, and Sven D{\"a}hne.
\newblock Learning how to explain neural networks: Patternnet and
  patternattribution.
\newblock \emph{arXiv preprint arXiv:1705.05598}, 2017.

\bibitem[Law et~al.(2019)Law, Snell, Farahmand, Urtasun, and
  Zemel]{law2019dimensionality}
Marc~T Law, Jake Snell, Amir-massoud Farahmand, Raquel Urtasun, and Richard~S
  Zemel.
\newblock Dimensionality reduction for representing the knowledge of
  probabilistic models.
\newblock In \emph{International Conference on Learning Representations}, 2019.

\bibitem[Le and Yang(2015)]{le2015tiny}
Ya~Le and Xuan Yang.
\newblock Tiny imagenet visual recognition challenge.
\newblock \emph{CS 231N}, 7:\penalty0 7, 2015.

\bibitem[Li et~al.(2020)Li, Zhao, and Scheidegger]{li2020visualizing}
Mingwei Li, Zhenge Zhao, and Carlos Scheidegger.
\newblock Visualizing neural networks with the grand tour.
\newblock \emph{Distill}, 2020.
\newblock \doi{10.23915/distill.00025}.
\newblock https://distill.pub/2020/grand-tour.

\bibitem[Li et~al.(2018)Li, Lu, Wang, Haupt, and Zhao]{li2018tighter}
Xingguo Li, Junwei Lu, Zhaoran Wang, Jarvis Haupt, and Tuo Zhao.
\newblock On tighter generalization bound for deep neural networks: Cnns,
  resnets, and beyond.
\newblock \emph{arXiv preprint arXiv:1806.05159}, 2018.

\bibitem[Liang et~al.(2019)Liang, Li, Li, and Zhang]{liang2019knowledge}
Ruofan Liang, Tianlin Li, Longfei Li, and Quanshi Zhang.
\newblock Knowledge consistency between neural networks and beyond.
\newblock In \emph{International Conference on Learning Representations}, 2019.

\bibitem[Lin et~al.(2014)Lin, Maire, Belongie, Hays, Perona, Ramanan,
  Doll{\'a}r, and Zitnick]{lin2014microsoft}
Tsung-Yi Lin, Michael Maire, Serge Belongie, James Hays, Pietro Perona, Deva
  Ramanan, Piotr Doll{\'a}r, and C~Lawrence Zitnick.
\newblock Microsoft coco: Common objects in context.
\newblock In \emph{European conference on computer vision}, pages 740--755.
  Springer, 2014.

\bibitem[Long and Sedghi(2020)]{long2019generalization}
Philip~M Long and Hanie Sedghi.
\newblock Generalization bounds for deep convolutional neural networks.
\newblock In \emph{International Conference on Learning Representations}, 2020.

\bibitem[Lundberg and Lee(2017)]{lunberg2017unified}
Scott~M Lundberg and Su-In Lee.
\newblock A unified approach to interpreting model predictions.
\newblock In I.~Guyon, U.~V. Luxburg, S.~Bengio, H.~Wallach, R.~Fergus,
  S.~Vishwanathan, and R.~Garnett, editors, \emph{Advances in Neural
  Information Processing Systems}, volume~30. Curran Associates, Inc., 2017.
\newblock URL
  \url{https://proceedings.neurips.cc/paper/2017/file/8a20a8621978632d76c43dfd28b67767-Paper.pdf}.

\bibitem[Madry et~al.(2017)Madry, Makelov, Schmidt, Tsipras, and
  Vladu]{madry2017towards}
Aleksander Madry, Aleksandar Makelov, Ludwig Schmidt, Dimitris Tsipras, and
  Adrian Vladu.
\newblock Towards deep learning models resistant to adversarial attacks.
\newblock \emph{arXiv preprint arXiv:1706.06083}, 2017.

\bibitem[Mahendran and Vedaldi(2015)]{mahendran2015understanding}
Aravindh Mahendran and Andrea Vedaldi.
\newblock Understanding deep image representations by inverting them.
\newblock In \emph{Proceedings of the IEEE conference on computer vision and
  pattern recognition}, pages 5188--5196, 2015.

\bibitem[Mardia and Jupp(2009)]{mardia2009directional}
Kanti~V Mardia and Peter~E Jupp.
\newblock \emph{Directional statistics}, volume 494.
\newblock John Wiley \& Sons, 2009.

\bibitem[Mordvintsev et~al.(2015)Mordvintsev, Olah, and
  Tyka]{google2015inceptionism}
Alexander Mordvintsev, Christopher Olah, and Mike Tyka.
\newblock Inceptionism: Going deeper into neural networks, 2015.
\newblock URL
  \url{https://research.googleblog.com/2015/06/inceptionism-going-deeper-into-neural.html}.

\bibitem[Neyshabur et~al.(2018)Neyshabur, Li, Bhojanapalli, LeCun, and
  Srebro]{neyshabur2018towards}
Behnam Neyshabur, Zhiyuan Li, Srinadh Bhojanapalli, Yann LeCun, and Nathan
  Srebro.
\newblock Towards understanding the role of over-parametrization in
  generalization of neural networks.
\newblock \emph{arXiv preprint arXiv:1805.12076}, 2018.

\bibitem[Novak et~al.(2018)Novak, Bahri, Abolafia, Pennington, and
  Sohl-Dickstein]{novak2018sensitivity}
Roman Novak, Yasaman Bahri, Daniel~A Abolafia, Jeffrey Pennington, and Jascha
  Sohl-Dickstein.
\newblock Sensitivity and generalization in neural networks: an empirical
  study.
\newblock \emph{arXiv preprint arXiv:1802.08760}, 2018.

\bibitem[Pearson(1901)]{pearson1901liii}
Karl Pearson.
\newblock Liii. on lines and planes of closest fit to systems of points in
  space.
\newblock \emph{The London, Edinburgh, and Dublin Philosophical Magazine and
  Journal of Science}, 2\penalty0 (11):\penalty0 559--572, 1901.

\bibitem[Radford et~al.(2021)Radford, Kim, Hallacy, Ramesh, Goh, Agarwal,
  Sastry, Askell, Mishkin, Clark, et~al.]{radford2021learning}
Alec Radford, Jong~Wook Kim, Chris Hallacy, Aditya Ramesh, Gabriel Goh,
  Sandhini Agarwal, Girish Sastry, Amanda Askell, Pamela Mishkin, Jack Clark,
  et~al.
\newblock Learning transferable visual models from natural language
  supervision.
\newblock \emph{arXiv preprint arXiv:2103.00020}, 2021.

\bibitem[Ren et~al.(2021{\natexlab{a}})Ren, Zhang, Wang, Chen, Zhou, Cheng,
  Wang, Chen, Shi, and Zhang]{ren2021advsarial}
Jie Ren, Die Zhang, Yisen Wang, Lu~Chen, Zhanpeng Zhou, Xu~Cheng, Xin Wang,
  Yiting Chen, Jie Shi, and Quanshi Zhang.
\newblock Game-theoretic understanding of adversarially learned features.
\newblock \emph{arXiv preprint arXiv:2103.07364}, 2021{\natexlab{a}}.

\bibitem[Ren et~al.(2021{\natexlab{b}})Ren, Zhou, Chen, and
  Zhang]{ren2021baseline}
Jie Ren, Zhanpeng Zhou, Qirui Chen, and Quanshi Zhang.
\newblock Learning baseline values for shapley values.
\newblock \emph{arXiv preprint arXiv:2105.10719}, 2021{\natexlab{b}}.

\bibitem[Ribeiro et~al.(2016)Ribeiro, Singh, and Guestrin]{ribeiro2016should}
Marco~T{\'{u}}lio Ribeiro, Sameer Singh, and Carlos Guestrin.
\newblock "why should {I} trust you?": Explaining the predictions of any
  classifier.
\newblock In \emph{Proceedings of the 22nd ACM SIGKDD international conference
  on knowledge discovery and data mining}, pages 1135--1144. ACM, 2016.

\bibitem[Roweis and Saul(2000)]{roweis2000nonlinear}
Sam~T Roweis and Lawrence~K Saul.
\newblock Nonlinear dimensionality reduction by locally linear embedding.
\newblock \emph{science}, 290\penalty0 (5500):\penalty0 2323--2326, 2000.

\bibitem[Sandler et~al.(2018)Sandler, Howard, Zhu, Zhmoginov, and
  Chen]{sandler2018mobilenetv2}
Mark Sandler, Andrew Howard, Menglong Zhu, Andrey Zhmoginov, and Liang-Chieh
  Chen.
\newblock Mobilenetv2: Inverted residuals and linear bottlenecks.
\newblock In \emph{Proceedings of the IEEE conference on computer vision and
  pattern recognition}, pages 4510--4520, 2018.

\bibitem[Selvaraju et~al.(2017)Selvaraju, Cogswell, Das, Vedantam, Parikh, and
  Batra]{selvaraju2017grad}
Ramprasaath~R Selvaraju, Michael Cogswell, Abhishek Das, Ramakrishna Vedantam,
  Devi Parikh, and Dhruv Batra.
\newblock Grad-cam: Visual explanations from deep networks via gradient-based
  localization.
\newblock In \emph{Proceedings of the IEEE international conference on computer
  vision}, pages 618--626, 2017.

\bibitem[Shapley(1953)]{shapley1953value}
Lloyd~S Shapley.
\newblock A value for n-person games.
\newblock \emph{Contributions to the Theory of Games}, 2\penalty0
  (28):\penalty0 307--317, 1953.

\bibitem[Shwartz-Ziv and Tishby(2017)]{shwartz2017opening}
Ravid Shwartz-Ziv and Naftali Tishby.
\newblock Opening the black box of deep neural networks via information.
\newblock \emph{arXiv preprint arXiv:1703.00810}, 2017.

\bibitem[Simonyan et~al.(2017)Simonyan, Vedaldi, and
  Zisserman]{simonyan2017deep}
K~Simonyan, A~Vedaldi, and A~Zisserman.
\newblock Deep inside convolutional networks: visualising image classification
  models and saliency maps.
\newblock \emph{arXiv preprint arXiv:1312.6034}, 2017.

\bibitem[Simonyan and Zisserman(2014)]{simonyan2014very}
Karen Simonyan and Andrew Zisserman.
\newblock Very deep convolutional networks for large-scale image recognition.
\newblock \emph{arXiv preprint arXiv:1409.1556}, 2014.

\bibitem[Simonyan et~al.(2013)Simonyan, Vedaldi, and
  Zisserman]{simonyan2013deep}
Karen Simonyan, Andrea Vedaldi, and Andrew Zisserman.
\newblock Deep inside convolutional networks: Visualising image classification
  models and saliency maps.
\newblock \emph{arXiv preprint arXiv:1312.6034}, 2013.

\bibitem[Sivic and Zisserman(2003)]{sivic2003video}
Josef Sivic and Andrew Zisserman.
\newblock Video google: A text retrieval approach to object matching in videos.
\newblock In \emph{IEEE International Conference on Computer Vision}, volume~3,
  pages 1470--1470. IEEE Computer Society, 2003.

\bibitem[Sra(2012)]{sra2012short}
Suvrit Sra.
\newblock A short note on parameter approximation for von mises-fisher
  distributions: and a fast implementation of i s (x).
\newblock \emph{Computational Statistics}, 27\penalty0 (1):\penalty0 177--190,
  2012.

\bibitem[Tenenbaum et~al.(2000)Tenenbaum, De~Silva, and
  Langford]{tenenbaum2000global}
Joshua~B Tenenbaum, Vin De~Silva, and John~C Langford.
\newblock A global geometric framework for nonlinear dimensionality reduction.
\newblock \emph{science}, 290\penalty0 (5500):\penalty0 2319--2323, 2000.

\bibitem[Tenney et~al.(2020)Tenney, Wexler, Bastings, Bolukbasi, Coenen,
  Gehrmann, Jiang, Pushkarna, Radebaugh, Reif, et~al.]{tenney2020language}
Ian Tenney, James Wexler, Jasmijn Bastings, Tolga Bolukbasi, Andy Coenen,
  Sebastian Gehrmann, Ellen Jiang, Mahima Pushkarna, Carey Radebaugh, Emily
  Reif, et~al.
\newblock The language interpretability tool: Extensible, interactive
  visualizations and analysis for nlp models.
\newblock \emph{arXiv preprint arXiv:2008.05122}, 2020.

\bibitem[Van~der Maaten and Hinton(2008)]{van2008visualizing}
Laurens Van~der Maaten and Geoffrey Hinton.
\newblock Visualizing data using t-sne.
\newblock \emph{Journal of machine learning research}, 9\penalty0 (11), 2008.

\bibitem[Wah et~al.(2011)Wah, Branson, Welinder, Perona, and
  Belongie]{WahCUB_200_2011}
C.~Wah, S.~Branson, P.~Welinder, P.~Perona, and S.~Belongie.
\newblock {The Caltech-UCSD Birds-200-2011 Dataset}.
\newblock Technical Report CNS-TR-2011-001, California Institute of Technology,
  2011.

\bibitem[Wang et~al.(2017{\natexlab{a}})Wang, Xiang, Cheng, and
  Yuille]{wang2017normface}
Feng Wang, Xiang Xiang, Jian Cheng, and Alan~Loddon Yuille.
\newblock Normface: L2 hypersphere embedding for face verification.
\newblock In \emph{Proceedings of the 25th ACM international conference on
  Multimedia}, pages 1041--1049, 2017{\natexlab{a}}.

\bibitem[Wang et~al.(2020{\natexlab{a}})Wang, Ren, Lin, Zhu, Wang, and
  Zhang]{wang2020transfer}
Xin Wang, Jie Ren, Shuyun Lin, Xiangming Zhu, Yisen Wang, and Quanshi Zhang.
\newblock A unified approach to interpreting and boosting adversarial
  transferability.
\newblock In \emph{International Conference on Learning Representations},
  2020{\natexlab{a}}.

\bibitem[Wang et~al.(2021)Wang, Lin, Zhang, Zhu, and
  Zhang]{wang2021adversarial}
Xin Wang, Shuyun Lin, Hao Zhang, Yufei Zhu, and Quanshi Zhang.
\newblock Interpreting attributions and interactions of adversarial attacks.
\newblock In \emph{Proceedings of the IEEE/CVF International Conference on
  Computer Vision}, pages 1095--1104, 2021.

\bibitem[Wang et~al.(2017{\natexlab{b}})Wang, Kucukelbir, and
  Blei]{wang2017robust}
Yixin Wang, Alp Kucukelbir, and David~M Blei.
\newblock Robust probabilistic modeling with bayesian data reweighting.
\newblock In \emph{International Conference on Machine Learning}, pages
  3646--3655. PMLR, 2017{\natexlab{b}}.

\bibitem[Wang et~al.(2020{\natexlab{b}})Wang, Turko, Shaikh, Park, Das, Hohman,
  Kahng, and Chau]{wang2020cnn}
Zijie~J Wang, Robert Turko, Omar Shaikh, Haekyu Park, Nilaksh Das, Fred Hohman,
  Minsuk Kahng, and Duen~Horng Chau.
\newblock Cnn explainer: Learning convolutional neural networks with
  interactive visualization.
\newblock \emph{IEEE Transactions on Visualization and Computer Graphics},
  2020{\natexlab{b}}.

\bibitem[Wen et~al.(2016)Wen, Zhang, Li, and Qiao]{wen2016discriminative}
Yandong Wen, Kaipeng Zhang, Zhifeng Li, and Yu~Qiao.
\newblock A discriminative feature learning approach for deep face recognition.
\newblock In \emph{European conference on computer vision}, pages 499--515.
  Springer, 2016.

\bibitem[Weng et~al.(2018)Weng, Zhang, Chen, Yi, Su, Gao, Hsieh, and
  Daniel]{weng2018evaluating}
Tsui-Wei Weng, Huan Zhang, Pin-Yu Chen, Jinfeng Yi, Dong Su, Yupeng Gao,
  Cho-Jui Hsieh, and Luca Daniel.
\newblock Evaluating the robustness of neural networks: An extreme value theory
  approach.
\newblock \emph{arXiv preprint arXiv:1801.10578}, 2018.

\bibitem[Williamson and Feng(2020)]{williamson2020efficient}
Brian~D Williamson and Jean Feng.
\newblock Efficient nonparametric statistical inference on population feature
  importance using shapley values.
\newblock \emph{In {ICML}}, 2020.

\bibitem[Wolchover(2017)]{wolchover2017new}
Natalie Wolchover.
\newblock New theory cracks open the black box of deep learning.
\newblock \emph{In {Quanta Magazine}}, 2017.

\bibitem[Xu and Raginsky(2017)]{xu2017information}
Aolin Xu and Maxim Raginsky.
\newblock Information-theoretic analysis of generalization capability of
  learning algorithms.
\newblock In \emph{Advances in Neural Information Processing Systems}, pages
  2524--2533, 2017.

\bibitem[Yosinski et~al.(2015)Yosinski, Clune, Nguyen, Fuchs, and
  Lipson]{yosinski2015understanding}
Jason Yosinski, Jeff Clune, Anh Nguyen, Thomas Fuchs, and Hod Lipson.
\newblock Understanding neural networks through deep visualization.
\newblock In \emph{International Conference on Machine Learning}, 2015.

\bibitem[Zeiler and Fergus(2014)]{zeiler2014visualizing}
Matthew~D Zeiler and Rob Fergus.
\newblock Visualizing and understanding convolutional networks.
\newblock In \emph{European conference on computer vision}, pages 818--833.
  Springer, 2014.

\bibitem[Zhang et~al.(2016)Zhang, Bengio, Hardt, Recht, and
  Vinyals]{zhang2016understanding}
Chiyuan Zhang, Samy Bengio, Moritz Hardt, Benjamin Recht, and Oriol Vinyals.
\newblock Understanding deep learning requires rethinking generalization.
\newblock \emph{arXiv preprint arXiv:1611.03530}, 2016.

\bibitem[Zhang et~al.(2019)Zhang, Bengio, and Singer]{zhang2019all}
Chiyuan Zhang, Samy Bengio, and Yoram Singer.
\newblock Are all layers created equal?
\newblock \emph{arXiv preprint arXiv:1902.01996}, 2019.

\bibitem[Zhang et~al.(2021{\natexlab{a}})Zhang, Zhou, Zhang, Bao, Huo, Chen,
  Cheng, Wu, and Zhang]{zhang2021nlptree}
Die Zhang, Huilin Zhou, Hao Zhang, Xiaoyi Bao, Da~Huo, Ruizhao Chen, Xu~Cheng,
  Mengyue Wu, and Quanshi Zhang.
\newblock Building interpretable interaction trees for deep nlp models.
\newblock In \emph{Proceedings of the AAAI Conference on Artificial
  Intelligence}, volume~35, pages 14328--14337, 2021{\natexlab{a}}.

\bibitem[Zhang et~al.(2020)Zhang, Li, Ma, Li, Xie, and Zhang]{zhang2020dropout}
Hao Zhang, Sen Li, YinChao Ma, Mingjie Li, Yichen Xie, and Quanshi Zhang.
\newblock Interpreting and boosting dropout from a game-theoretic view.
\newblock In \emph{International Conference on Learning Representations}, 2020.

\bibitem[Zhang et~al.(2021{\natexlab{b}})Zhang, Xie, Zheng, Zhang, and
  Zhang]{zhang2021multivariate}
Hao Zhang, Yichen Xie, Longjie Zheng, Die Zhang, and Quanshi Zhang.
\newblock Interpreting multivariate shapley interactions in dnns.
\newblock In \emph{Proceedings of the AAAI Conference on Artificial
  Intelligence}, volume~35, pages 10877--10886, 2021{\natexlab{b}}.

\bibitem[Zhou et~al.(2015)Zhou, Khosla, Lapedriza, Oliva, and
  Torralba]{zhou2014object}
Bolei Zhou, Aditya Khosla, Agata Lapedriza, Aude Oliva, and Antonio Torralba.
\newblock Object detectors emerge in deep scene cnns.
\newblock In \emph{International Conference on Learning Representations}, 2015.

\bibitem[Zhou et~al.(2016)Zhou, Khosla, Lapedriza, Oliva, and
  Torralba]{zhou2016learning}
Bolei Zhou, Aditya Khosla, Agata Lapedriza, Aude Oliva, and Antonio Torralba.
\newblock Learning deep features for discriminative localization.
\newblock In \emph{Proceedings of the IEEE conference on computer vision and
  pattern recognition}, pages 2921--2929, 2016.

\bibitem[Zintgraf et~al.(2017)Zintgraf, Cohen, Adel, and
  Welling]{zintgraf2017visualizing}
Luisa~M Zintgraf, Taco~S Cohen, Tameem Adel, and Max Welling.
\newblock Visualizing deep neural network decisions: Prediction difference
  analysis.
\newblock In \emph{International Conference on Learning Representations}, 2017.

\end{thebibliography}

\newpage
\appendix

\vspace{-1.5em}
\section{More experimental results}
\label{sec:more-result}

This section shows more experimental results based on different DNNs and datasets.
Besides, we have also included a demo video at {\small\url{https://youtu.be/bnbVw2vBVQ8}} to better illustrate the temporal-spatial emergence of discriminative regional patterns in Figure 3 and Figure 4 of the paper.

\subsection{More visualization of the sample-wise discrimination power}
\begin{figure}[htbp]
    \centering
    \includegraphics[width=\linewidth]{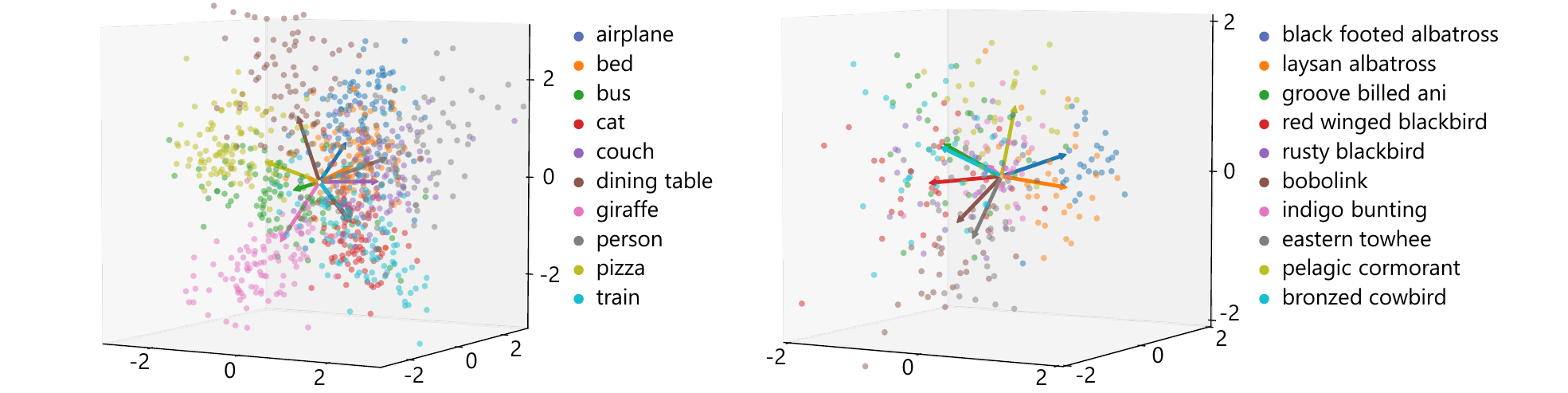}
    \caption{{\small Visualization of sample features learned by (left) ResNet-50 in the coordinate system of visualizing the COCO 2014 categories, and by (right) ResNet-34 in the coordinate system of visualizing the CUB-200-2011 categories\textsuperscript{\ref{fn:detail}}.}}
    \vspace{-1em}
    \label{fig:s-emb}
\end{figure}

\begin{figure}[htbp]
    \centering
    \includegraphics[width=\linewidth]{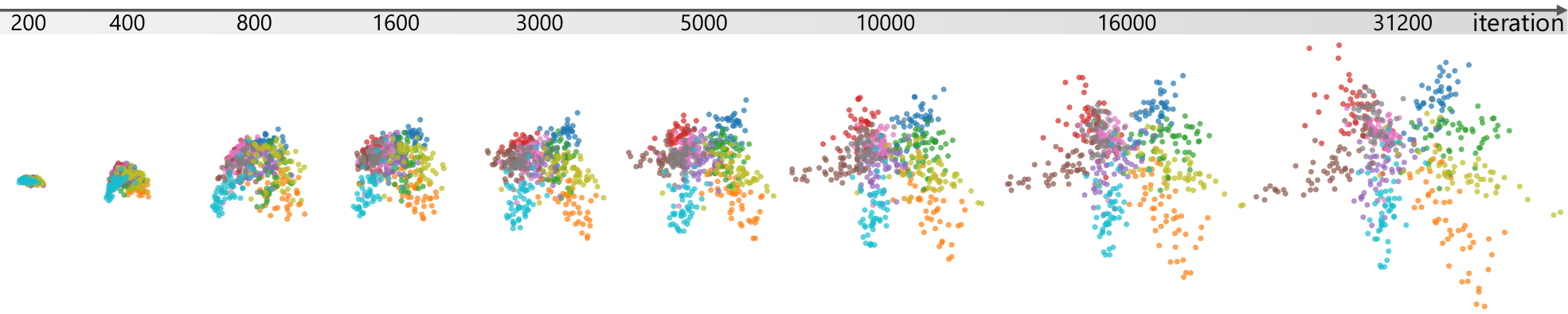}
    \caption{{\small The increasing discrimination power of sample features in VGG-16 during the training process in the coordinate system of visualizing the Tiny ImageNet categories\textsuperscript{\ref{fn:detail}}.}}
    \label{fig:s-emb-temporal}
\end{figure}

Figure \ref{fig:s-emb} shows the projected sample features {\small$g$} and each category direction {\small$\mu_c$}, based on the COCO 2014 dataset and the CUB-200-2011 dataset\footnote{\label{fn:detail}Please see Section \ref{sec:exp-detail} for details of the dataset, and the selection of sample features and regional features.}.
The visualization results revealed the semantic similarity between categories.
For example, \textit{dining table} features were similar to \textit{pizza} features, and \textit{black footed albatross} features were similar to \textit{laysan albatross} features.
Furthermore, Figure \ref{fig:s-emb-temporal} shows the projected sample feature {\small$g$} at different iterations of training.
This illustrated how the discrimination power of sample features gradually increased through the training process.

\begin{figure}[htbp]
    \centering
    \includegraphics[width=\linewidth]{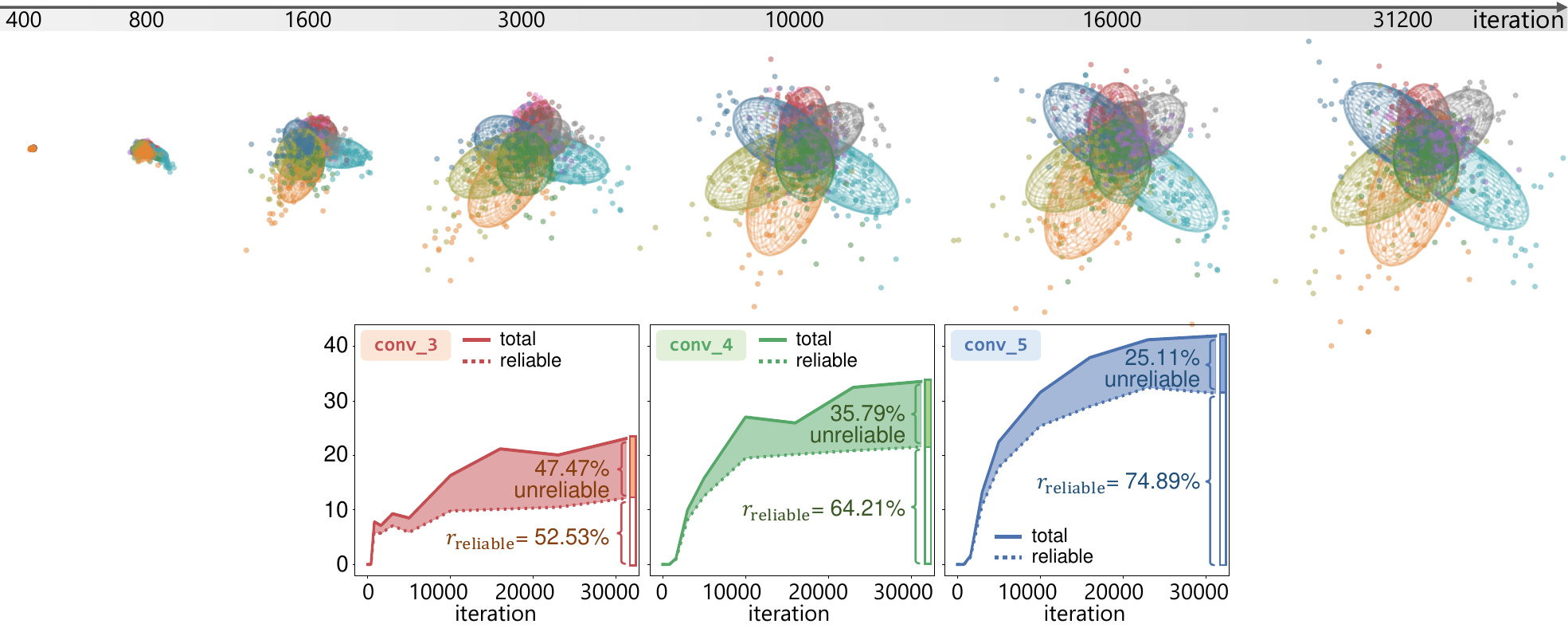}
    \vspace{-.8em}
    \caption{{\small (top) The emergence of regional patterns in ResNet-34 through the learning process in the coordinate system of visualizing the Tiny ImageNet categories\textsuperscript{\ref{fn:detail}}. (bottom) The increase of total knowledge points and reliable knowledge points during training. The ratio of reliable knowledge points, {\small$r_{\text{reliable}}\!=\!{\text{\# of reliable points}/\text{\# of all points}}$}, increases through the forward propagation.}}
    \label{fig:r-emb-temporal}
\end{figure}

\begin{figure}[htbp]
    \centering
    \subfigure[\label{fig:r-emb-spatial-rn34-tiny}The emergence of regional patterns in ResNet-34 trained on the Tiny ImageNet dataset.]{
        \includegraphics[width=\linewidth]{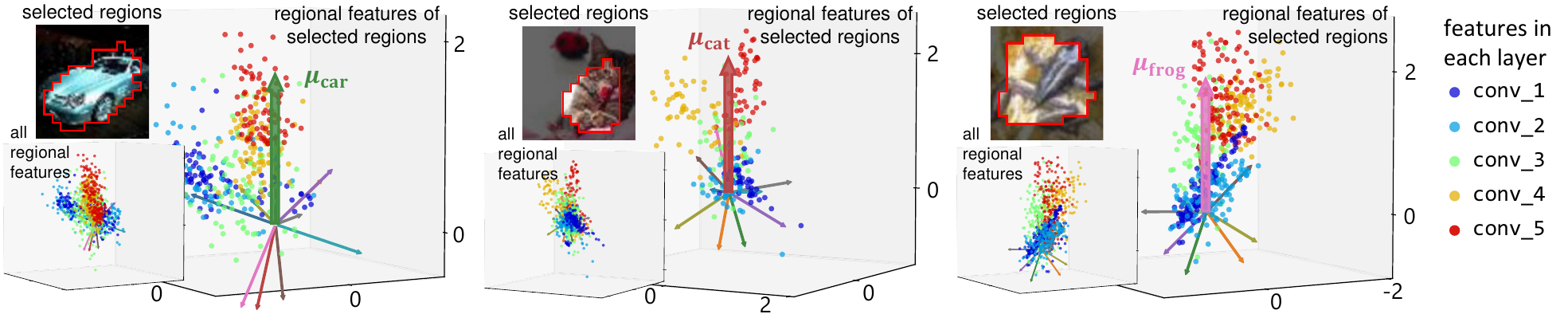}
    }
    \subfigure[\label{fig:r-emb-spatial-mnet-tiny}The emergence of regional patterns in MobileNet-V2 trained on the Tiny ImageNet dataset.]{
        \includegraphics[width=\linewidth]{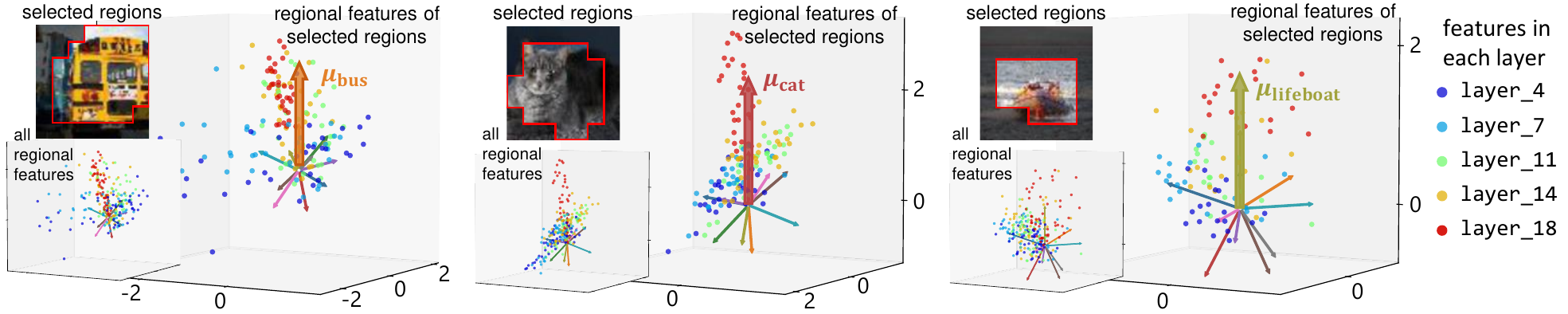}
    }
    \subfigure[\label{fig:r-emb-spatial-rn50-coco}The emergence of regional patterns in ResNet-50 trained on the COCO dataset.]{
        \includegraphics[width=\linewidth]{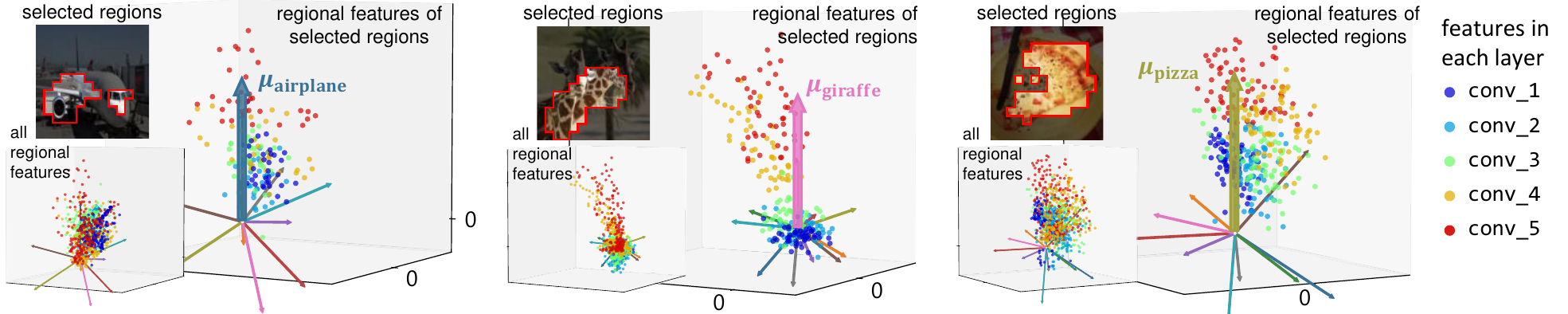}
    }
    \subfigure[\label{fig:r-emb-spatial-rn34-cub}The emergence of regional patterns in ResNet-34 trained on the CUB dataset.]{
        \includegraphics[width=\linewidth]{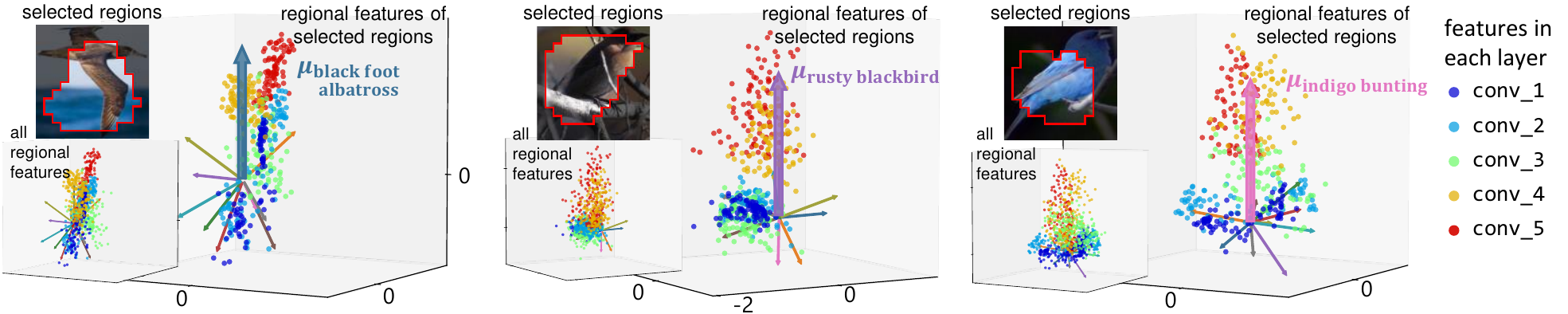}
    }
    \vspace{-.8em}
    \caption{The emergence of regional patterns through the forward propagation. Coordinates along the vertical axis reflect the discrimination power of the target category.}
    \label{fig:r-emb-spatial}
\end{figure}

\subsection{More visualization of the regional discrimination power}
Figure \ref{fig:r-emb-temporal}(top) illustrates the increasing discrimination power of ResNet-34 regional features\textsuperscript{\ref{fn:detail}} {\small$h^{(r)}$} during the training process, which was trained on the Tiny ImageNet dataset.
Figure \ref{fig:r-emb-temporal}(bottom) also shows the increase of knowledge points and reliable knowledge points in different layers during the training process.
In addition, we found that the ratio of reliable knowledge points in high layers (\emph{e.g.} the {\small\texttt{conv\_5}} layer\textsuperscript{\ref{fn:detail}}) was higher than that in low layers (\emph{e.g.} the {\small\texttt{conv\_3}} layer\textsuperscript{\ref{fn:detail}}).
This demonstrated that the DNN used non-discriminative and unreliable patterns in low layers to construct discriminative and reliable patterns in high layers.
Furthermore, Figure \ref{fig:r-emb-spatial} shows the emergence of discriminative regional features during the forward propagation, as a supplement to Figure 4 in the paper.

\subsection{More visualization of knowledge points}
\begin{figure}[htbp]
    \centering
    \subfigure[\label{concept-rn34-tiny}Image regions of knowledge points in ResNet-34 learned on the Tiny ImageNet dataset.]{
        \includegraphics[width=\linewidth]{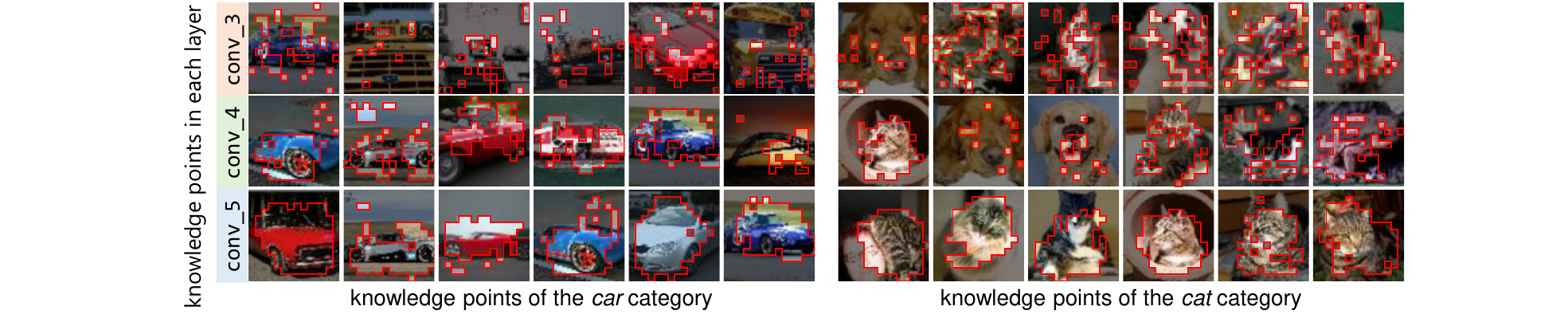}
    }
    \subfigure[\label{concept-rn50-coco}Image regions of knowledge points in ResNet-50 learned on the COCO 2014 dataset.]{
        \includegraphics[width=\linewidth]{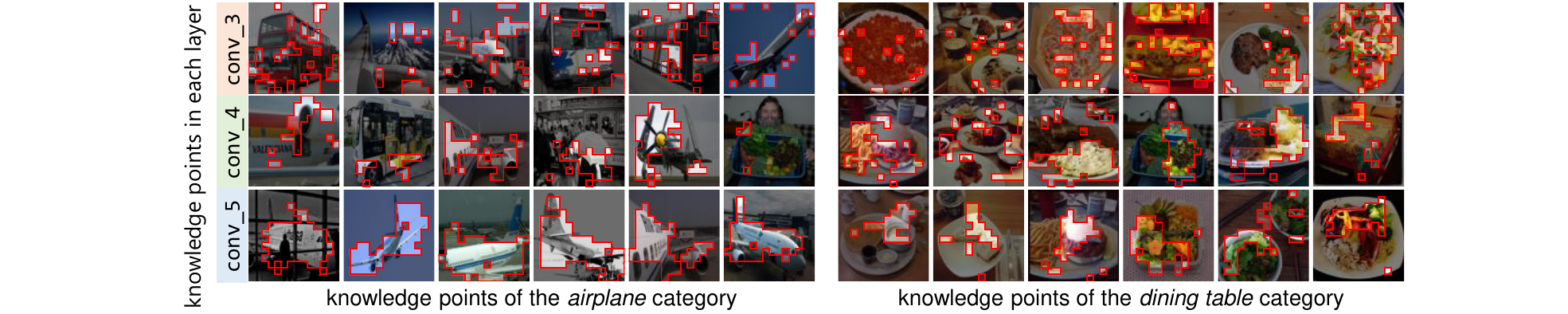}
    }
    \caption{{\small Visualization of image regions corresponding to knowledge points towards different categories.}}
    \label{fig:concept}
\end{figure}

Similar to Figure 6(right) in the paper, Figure \ref{fig:concept} highlights the image regions corresponding to knowledge points in different layers of the DNN.
Regional features in high layers were usually more likely to be localized on the foreground than regional features in low layers.

\subsection{More visualization about the adversarial attack}

\begin{figure}[htbp]
    \centering
    \includegraphics[width=\linewidth]{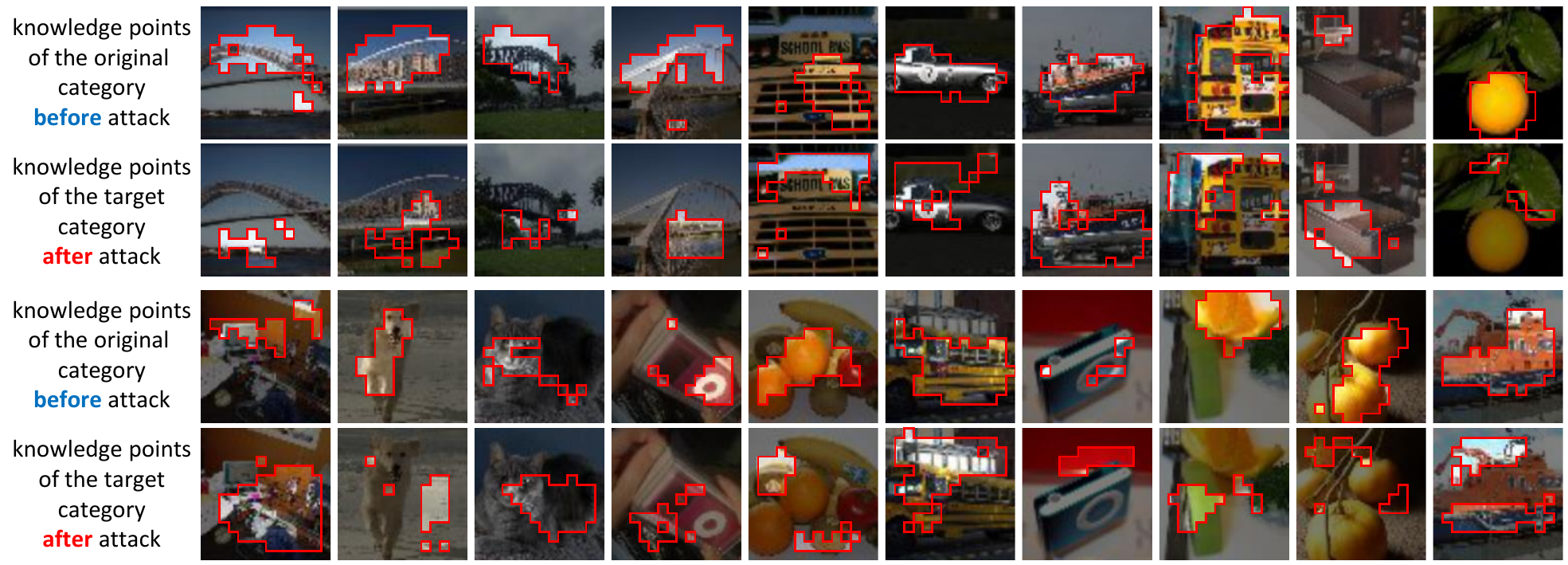}
    \caption{{\small Image regions corresponding to knowledge points in original and adversarial samples.}}
    \label{fig:adversarial}
\end{figure}

As a supplement to Figure 7(left) in the paper, Figure \ref{fig:adversarial} visualizes more image regions corresponding to knowledge points in original and adversarial samples.
We found that important regions for the classification of the original image (the first and third row) were usually different from important regions that attacked the classification towards the target category (the second and fourth row).
It means that the adversarial attack usually forces unimportant regional features in the original category to become important features in the target category.

\subsection{More verification of the estimated regional importance}

\begin{figure}
    \centering
    \subfigure[The estimated regional importance {\small$w^{(r)}$} and Shapley values {\small$\phi^{(r)}$} of \texttt{conv\_53} features in VGG-16 trained on the Tiny ImageNet dataset.]{\includegraphics[width=.95\linewidth]{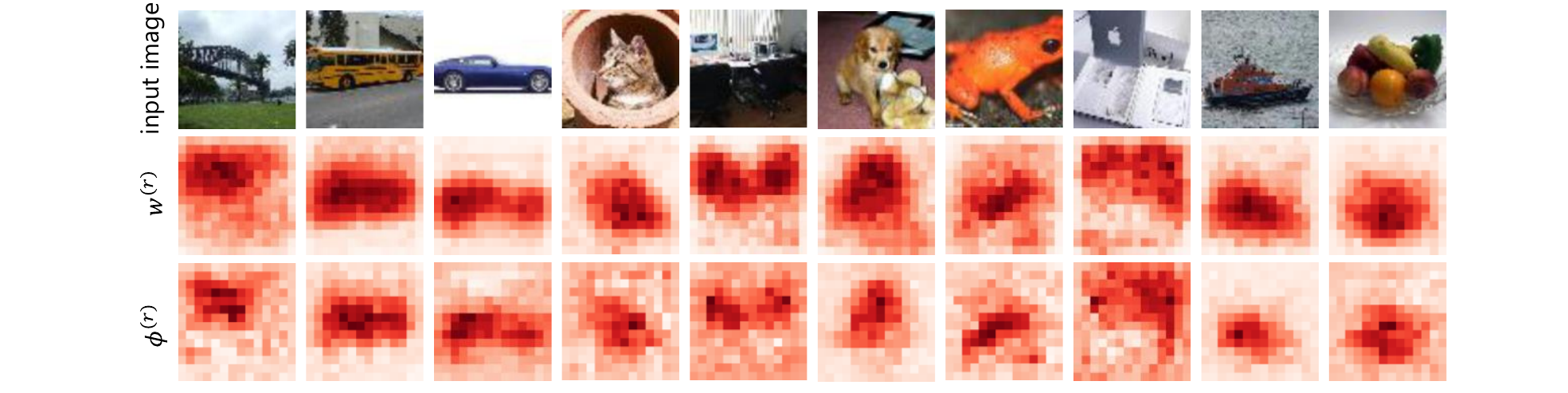}}
    \subfigure[The estimated regional importance {\small$w^{(r)}$} and Shapley values {\small$\phi^{(r)}$} of \texttt{conv\_5} features\textsuperscript{\ref{fn:detail}} in ResNet-34 trained on the Tiny ImageNet dataset.]{\includegraphics[width=.95\linewidth]{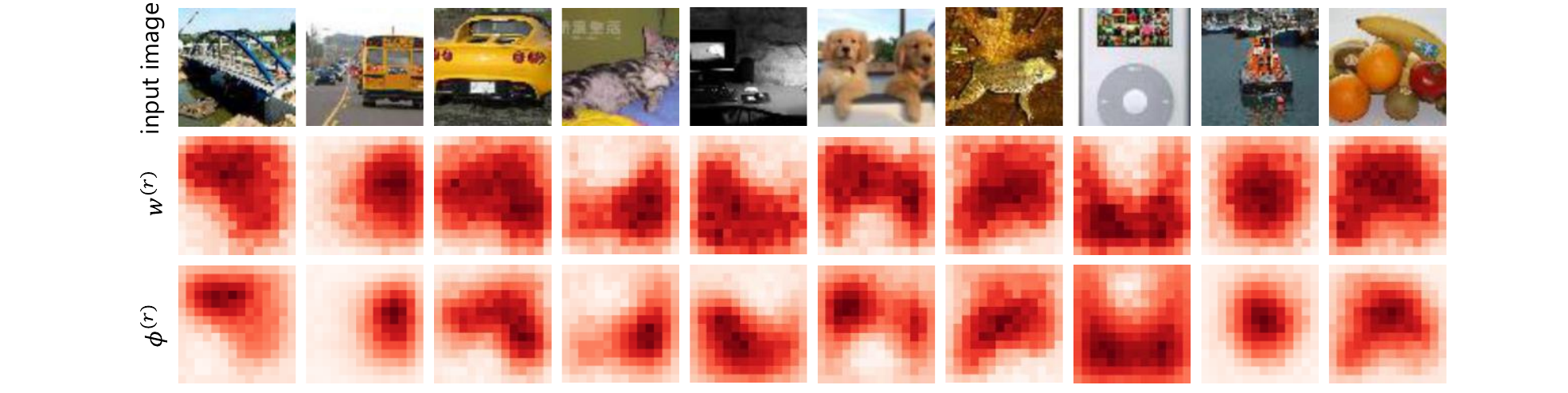}}
    \subfigure[The estimated regional importance {\small$w^{(r)}$} and Shapley values {\small$\phi^{(r)}$} of \texttt{layer\_18} features\textsuperscript{\ref{fn:detail}} in MobileNet-V2 trained on the Tiny ImageNet dataset.]{\includegraphics[width=.95\linewidth]{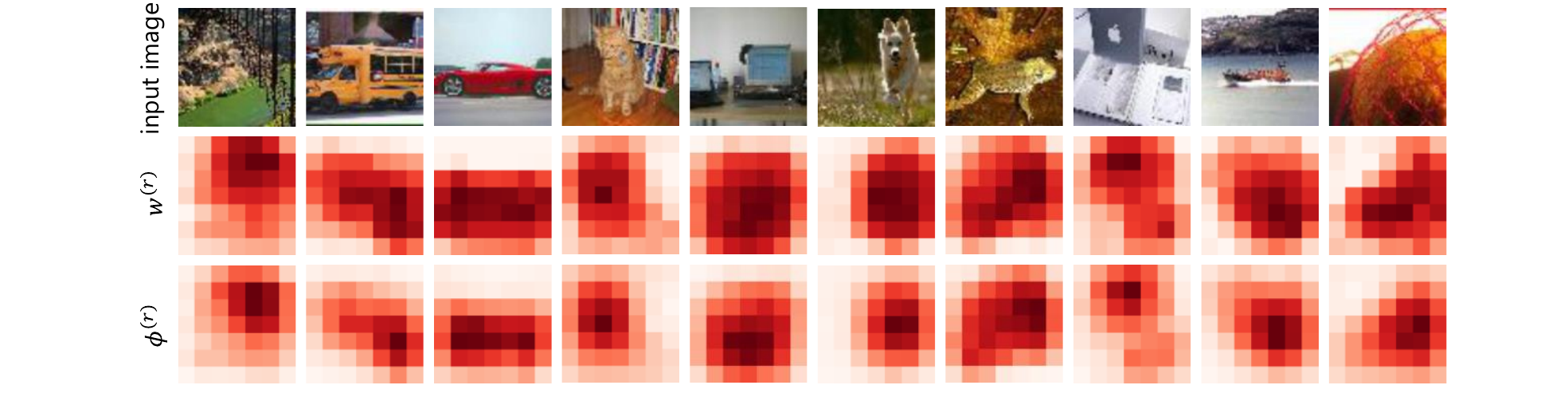}}
    \subfigure[The estimated regional importance {\small$w^{(r)}$} and Shapley values {\small$\phi^{(r)}$} of \texttt{conv\_5} features\textsuperscript{\ref{fn:detail}} in ResNet-50 trained on the COCO 2014 dataset.]{\includegraphics[width=.95\linewidth]{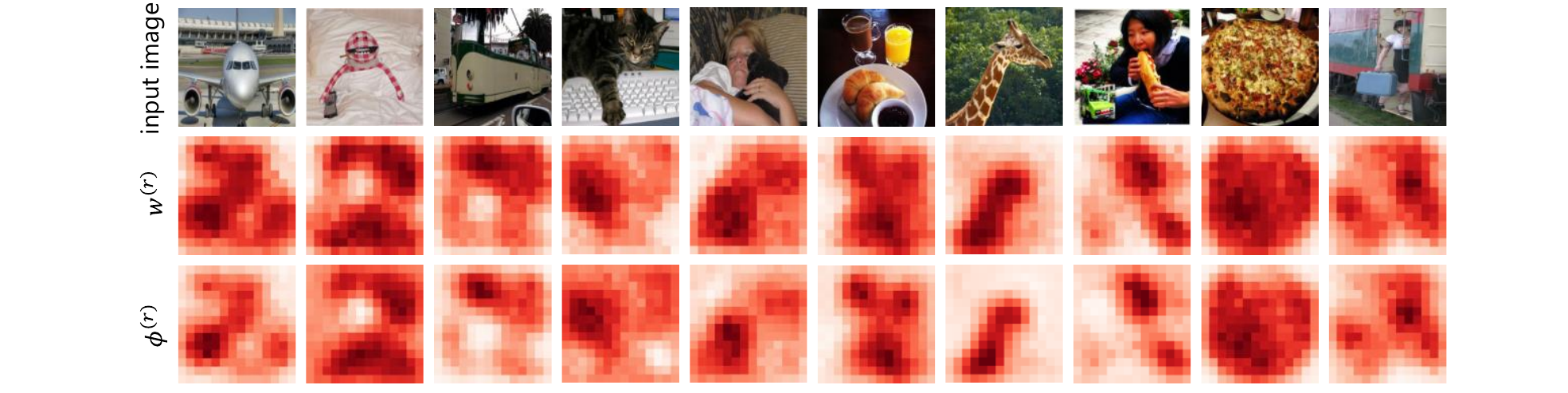}}
    \subfigure[The estimated regional importance {\small$w^{(r)}$} and Shapley values {\small$\phi^{(r)}$} of \texttt{conv\_5} features\textsuperscript{\ref{fn:detail}} in ResNet-34 trained on the CUB dataset.]{\includegraphics[width=.95\linewidth]{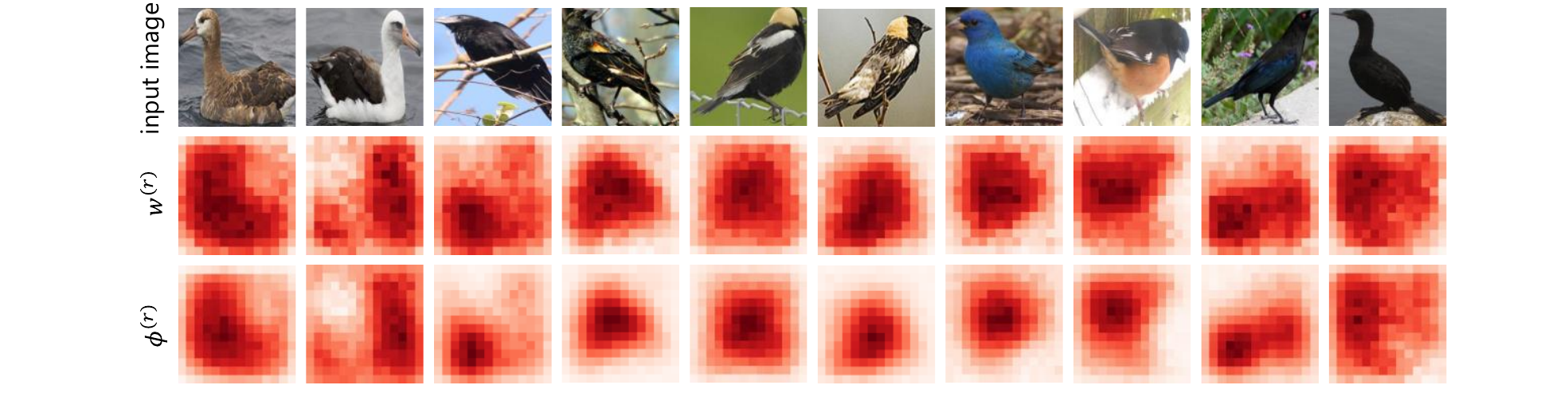}}
    \caption{{\small Visualization of the estimated regional importance and the regional Shapley value. Our regional importance is similar to the Shapley value of {\small$f^{(r)}$}, which verifies the trustworthiness of our method.}}
    \label{fig:w-shap}
\end{figure}

In this section, we provide more comparisons between the regional importance {\small$w^{(r)}$} and the Shapley values {\small$\phi^{(r)}$} based on DNNs trained on other datasets, as a supplement to Figure 5 in the paper.
Figure \ref{fig:w-shap} shows the high similarity between {\small$w^{(r)}$} and {\small$\phi^{(r)}$} among different regions {\small$r$}, based on various DNNs and datasets.
This demonstrated the trustworthiness of the estimated regional importance.

\subsection{The demo video to illustrate the temporal-spatial emergence of regional patterns}

We also provide a demo video at {\small\url{https://youtu.be/bnbVw2vBVQ8}} to better illustrate the emergence of discriminative regional patterns in both temporal and spatial manner.
Specifically, the upper half of the video demonstrates the trajectory of projected regional features {\small$h^{(r)}$} in the {\small\texttt{conv\_53}} layer of VGG-16 during the training process, with a progress bar showing the number of iterations at the bottom.
In addition, the lower half shows four examples of {\small$h^{(r)}$}'s trajectory through the forward propagation in four different samples.
We also provide a progress bar showing names of layers at the bottom.
For better illustration, we used the B-spline interpolation method~\cite{dierckx1995curve} to smooth the trajectory of regional features through the training process and during the forward propagation.
These demo videos can help people better understand the temporal-spatial emergence of discriminative regional features in a DNN.

\section{Discussions about the revised vMF distribution}
\label{sec:discuss-vmf}

\textbf{The vMF distribution} is a kind of distribution for modeling data on a sphere~\cite{fisher1953dispersion,fisher1993statistical,mardia2009directional}.
It is one of the simplest distributions for directional data.
Specifically, a vMF distribution on the {\small$(d-1)$}-sphere {\small$\mathbb{S}^{d-1}$} in {\small$\mathbb{R}^d$} is parameterized by the mean direction {\small$\mu\in\mathbb{R}^d$}, and the concentration parameter {\small$\kappa\geq0$}.
Suppose {\small$f\in\mathbb{R}^d$} follows the vMF distribution.
Then, the probability density function of {\small$f$} is given by
\begin{small}
\begin{equation}
    p_{\text{vMF}}(f|\mu,\kappa)=C_d(\kappa)\cdot\exp[\kappa\cdot\cos(\mu,f)],
\end{equation}
\end{small}
where {\small$C_d(\kappa)=\frac{\kappa^{d/2-1}}{(2\pi)^{d/2}I_{d/2-1}(\kappa)}$} is the normalization constant and {\small$I_{d/2-1}(\cdot)$} denotes the modified Bessel function of the first kind at order {\small$d/2-1$}~\cite{abramowitz1972handbook}.
Actually, the vMF distribution can be considered as a spherical analogue to the Gaussian distribution on the unit sphere.
{\small$\mu$} measures the mean direction.
{\small$\kappa$} controls the variance of {\small$f$}'s orientation \emph{w.r.t.} the mean direction {\small$\mu$}. A large value of {\small$\kappa$} implies a low variance \emph{w.r.t.} {\small$\mu$}.
In particular, when {\small$\kappa=0$}, the distribution reduces to a uniform distribution on {\small$\mathbb{S}^{d-1}$}; when {\small$\kappa\to\infty$}, the distribution reduces to a point density.
In this way, the probability density of {\small$f$} only depends on its orientation.

\textbf{The revised vMF distribution} mentioned in Section 3.1 of the paper takes into account the noise in {\small$f$}, \emph{i.e.} {\small$f=f^\star+\epsilon$}, {\small$\epsilon\sim\mathcal{N}(\bm{0},\sigma^2I_{d})$}.
In this case, all features {\small$f$} of a specific strength {\small$l=\Vert f\Vert_2$} have similar vulnerabilities to noises.
Features of different strengths have different vulnerabilities to noises.
Therefore, the probability density of {\small$f$} not only depends on its orientation but also its strength.
In Eq. (3) of the paper, we assume that all features {\small$f$} of a specific strength {\small$l$} follow a vMF distribution with a specific {\small$\kappa(l)$}.
The concentration parameter {\small$\kappa(l)$} is determined based on statistics of all features of the same strength {\small$l$}.
Specifically, to quantify {\small$\kappa(l)$}, we first sample {\small$\{f^\star_i\}_{i=1}^N$} from {\small$p_{\text{vMF}}(\mu,\kappa)$}.
Then, the noise {\small$\epsilon\sim\mathcal{N}(\bm{0},\sigma^2I_{d})$} is added to each sample {\small$f^\star_i$}, \emph{i.e.} {\small$f_i=f^\star_i+\epsilon$}.
Since we assume that {\small$f_i$} also follows a vMF distribution, we can estimate {\small$\kappa(l)$} via maximum likelihood estimation (MLE)~\cite{sra2012short}, as follows.
\begin{small}
\begin{equation}
    \kappa(l)=\mathop{\arg\max}_{\hat{\kappa}}\prod_{i=1}^{N} p_{\text{vMF}}(f_i|\mu,\hat{\kappa})
    \ \Rightarrow\
    \kappa(l)=\frac{\Vert \bar{f}\Vert_2(d-\Vert \bar{f}\Vert_2^2)}{1-\Vert \bar{f}\Vert_2^2},
\end{equation}
\end{small}
where {\small$\bar{f}=\frac{1}{N}\sum_{i}f_i/\Vert f_i\Vert_2$}.
In the calculation of {\small$\kappa(l)$}, the sample number {\small$N$} was set to 10000, and {\small$\sigma$} was set to 1.

\section{Derivations on the learning of the mixture model in sample feature visualization via the EM algorithm}
\label{sec:derivation-s-emb}

This section provides detailed derivations on the learning of the mixture model in Section 3.2 of the paper.
In the learning of mixture-model parameters {\small$\{\pi,\mu\}=\{\pi_y,\mu_y\}_{y\in Y}$}, we used the EM algorithm to maximize the likelihood {\small$\max_{\{\pi,\mu\}}\prod_{g}p(g)$}.
In this way, {\small$\{\pi,\mu\}$} were updated via the E-step and the M-step.
\begin{small}
\begin{equation}
\begin{aligned}
\text{(E-step)}
\qquad &
p(y|g)=\frac{\pi_y\cdot\exp\left[\kappa(l_g)\cdot\cos(o_g,\mu_y)\right]}{\sum_{y'}\pi_{y'}\cdot\exp\left[\kappa(l_g)\cdot\cos(o_g,\mu_{y'})\right]}\\[3pt]
\text{(M-step)}
\qquad &
\mu_y\propto\mathbb{E}\left[\kappa(l_g)\cdot p(y|g)\cdot o_g\right]_{\text{given }x},\
\pi_y=\mathbb{E}_x[p(y|g)]_{\text{given }x},
\end{aligned}
\end{equation}
\end{small}
where {\small$l_g=\Vert g\Vert_2$} and {\small$o_g=g/l_g$} denote the strength and orientation of {\small$g$}.
The derivation is similar to that in~\cite{banerjee2005clusteringbregman,banerjee2005clusteringvmf}.

\section{Additional verification of the effectiveness of sample-feature visualization}
\label{sec:verification-s-emb}

\begin{figure}[htbp]
    \centering
    \vspace{-.3em}
    \includegraphics[width=\linewidth]{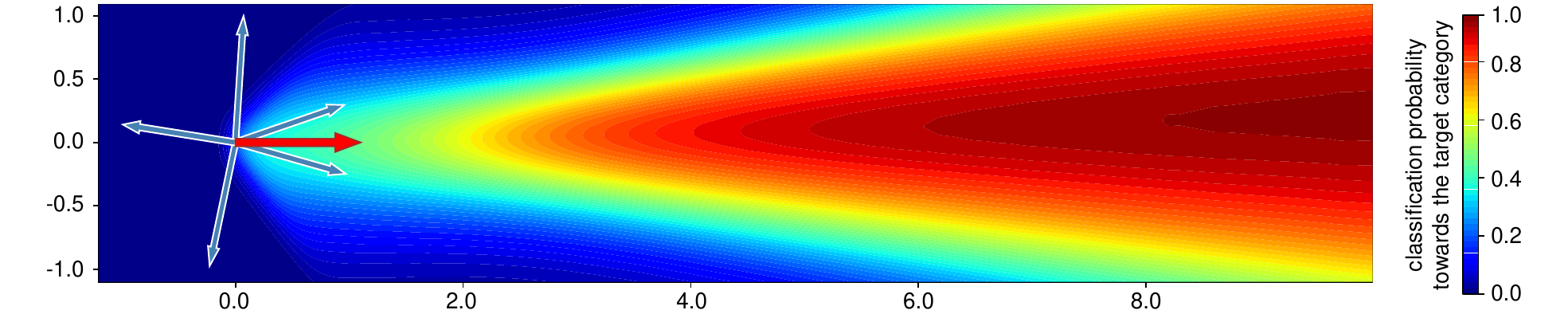}
    \vspace{-1em}
    \caption{{\small Classification probability towards the target category. The mean direction of the target category is illustrated as the red arrow. We found that sample features {\small$g$} with large strength were usually more confident towards classification.}}
    \label{fig:contour}
    \vspace{-.5em}
\end{figure}

In this section, we further verify the effectiveness of sample-feature visualization by showing a contour map of the classification probability of the sample feature {\small$g$}.
In Figure \ref{fig:contour}, we consider a toy example for the classification of six classes.
The red arrow represents the mean direction of the target category, while blue arrows are mean directions of other categories.
Figure \ref{fig:contour} shows the classification probability towards the target category.
We found that sample features {\small$g$} with large strength were more confident towards classification, which further verified the conclusion in Paragraph \textit{visualization and verification of sample features' discrimination power}, Section 4.

\section{Derivations of the equivalent form of the loss $\mathcal{L}_{\text{align}}$}
\label{sec:derivation-r-emb}
This section gives detailed derivations of {\small$\mathcal{L}_{\text{align}}=-\mathbb{E}_x[\sum_r w^{(r)}\cdot\cos(g,h^{(r)})]$} in Section 3.3 of the paper.
According to Eq. (10) in the paper, the optimization of {\small$\mathcal{L}_{\text{align}}$} can be written as follows.
\begin{small}
\begin{equation}
\frac{\partial \mathcal{L}_{\text{align}}}{\partial\Lambda} = \frac{\partial}{\partial\Lambda}\left[\mathbb{E}_{Q_\Lambda(\bm{h})}\left(\log Q_\Lambda(\bm{h})\right)-\mathbb{E}_{Q_\Lambda(\bm{h},g)}\left(\log Q_\Lambda(\bm{h}|g)\right)\right],
\end{equation}
\end{small}
where {\small$Q_\Lambda(\bm{h})$} is the prior of regional features {\small$\bm{h}$}.
For simplicity, we treat {\small$Q_\Lambda(\bm{h})$} as a constant.
Therefore, the optimization can be derived as follows.
\begin{small}
\begin{equation}
\begin{aligned}
\frac{\partial \mathcal{L}_{\text{align}}}{\partial\Lambda} &= -\mathbb{E}_{Q_\Lambda(\bm{h},g)}\left[\frac{\partial\log Q_\Lambda(\bm{h}|g)}{\partial\Lambda}\right]\\
&= -\mathbb{E}_x\left[{\sum}_r w^{(r)}\cdot \frac{\partial\log Q_\Lambda(h^{(r)}|g)}{\partial\Lambda}\right]_{\text{given }x}\\
&= -\mathbb{E}_x\left[{\sum}_r w^{(r)}\cdot \frac{\partial\log p_{\text{vMF}}(h^{(r)}|\mu=g,\kappa')}{\partial\Lambda}\right]_{\text{given }x}\\
&= -\mathbb{E}_x\left[{\sum}_r w^{(r)}\cdot \frac{\partial\log \frac{1}{C_{d'}(\kappa')}\exp[\kappa'\cdot\cos(g,h^{(r)})]}{\partial\Lambda}\right]_{\text{given }x}\\
&= -\mathbb{E}_x\left[{\sum}_r w^{(r)}\cdot \frac{\partial \kappa'\cdot\cos(g,h^{(r)})}{\partial\Lambda}\right]_{\text{given }x}\\
&= -\kappa'\cdot\frac{\partial}{\partial\Lambda}\mathbb{E}_x\left[{\sum}_r w^{(r)}\cdot \cos(g,h^{(r)})\right]_{\text{given }x}
\end{aligned}
\end{equation}
\end{small}
{\small$\kappa'$} is a positive constant, which does not essentially affect the convergence of {\small$\Lambda$}.
Therefore, the loss {\small$\mathcal{L}_{\text{align}}$} can be equivalently written as {\small$\mathcal{L}_{\text{align}}=-\mathbb{E}_x[\sum_r w^{(r)}\cdot\cos(g,h^{(r)})]$}.

\section{Discussions about the quantification of knowledge points}
\label{sec:discuss-kp}

This section provides more discussions on the quantification of knowledge points.
According to Section 3.4 of the paper, a regional feature is a knowledge point if it is discriminative enough for classification, \emph{i.e.} {\small$\max_c p(y=c|h^{(r)})>\tau$}.
Actually, there is a trade off between the value of {\small$\tau$} and the number of knowledge points.
If the value of {\small$\tau$} is large, then only a few regional features that are discriminative enough will be quantified as knowledge points.
On the other hand, if the value of {\small$\tau$} is small, then a large number of regional features will be quantified as knowledge points. Some of them are not so discriminative.
Therefore, we chose {\small$\tau=0.4$} to balance the trade-off between the discrimination power and the number of knowledge points.

Besides, setting the same value of {\small$\tau$} enables fair comparisons of the discrimination power between features in different layers.
First, for each layer, all the {\small$HW$} regional features in {\small$\bm{h}$} were learned to mimic the sample-wise distribution {\small$P(x_2|x_1)$} inferred by the DNN.
Second, for each layer, we uniformly sampled and analyzed the same number of regions.
In this way, for each layer, our method used the same number of regional features to mimic the same sample-wise distribution {\small$P(x_2|x_1)$}, making the learned regional feature {\small$h^{(r)}$} fairly represent the relative discrimination power of each region, and enabling fair comparisons between regional features through different layers.
Furthermore, when quantifying knowledge points in different layers of the DNN, we also normalized the average strength of regional features {\small$\mathbb{E}_{x,r}[\Vert h^{(r)}\Vert_2\ _{\text{given }x}]$} in each layer to the same value.
This also ensures the fair comparison between regional features in each layer.

\section{Settings of additional experiments in the supplementary material}
\label{sec:exp-detail}

\textbf{Datasets.} We conducted experiments on the task of object classification using the Tiny ImageNet dataset~\cite{le2015tiny}, the MS COCO 2014 dataset~\cite{lin2014microsoft}, and the CUB-200-2011 dataset~\cite{WahCUB_200_2011}.
To clarify the visualization result, we randomly selected ten categories from each dataset.
For the Tiny ImageNet dataset, we selected \textit{steel arch bridge (bridge), school bus (bus), sports car (car), tabby cat (cat), desk, golden retriever (dog), tailed frog (frog), iPod, lifeboat}, and \textit{orange} for classification.
For the MS COCO 2014 dataset, we selected \textit{airplane, bed, bus, cat, couch, dining table, giraffe, person, pizza}, and \textit{train} for classification.
For the CUB-200-2011 dataset, we selected \textit{black footed albatross, laysan albatross, groove billed ani, red winged blackbird, rusty blackbird, bobolink, indigo bunting, eastern towhee, pelagic cormorant}, and \textit{bronzed cowbird} for classification.
We used images cropped by the annotated bounding boxes in the MS COCO 2014 dataset and the CUB-200-2011 dataset.

\textbf{DNNs, and the selection of sample features and regional features.}
We analyzed intermediate-layer features in VGG-16~\cite{simonyan2014very}, ResNet-34/50~\cite{he2016deep}, MobileNet-V2~\cite{sandler2018mobilenetv2}.
We slightly modified the ResNets by changing the stride in {\small\texttt{conv\_5x}} layers to 1.
For each of the DNNs, we used the feature before the last fully-connected layer as the raw sample feature.
We analyzed regional features in different layers for each DNN.
For the VGG-16, we selected the output feature of the {\small\texttt{conv\_12}}, {\small\texttt{conv\_22}}, {\small\texttt{conv\_33}}, {\small\texttt{conv\_43}}, and {\small\texttt{conv\_53}} layers as the raw regional feature.
For ResNets, we selected the output feature of the {\small\texttt{conv\_1}}, {\small\texttt{conv\_2x}}, {\small\texttt{conv\_3x}}, {\small\texttt{conv\_4x}}, and {\small\texttt{conv\_5x}} layers as the raw regional feature (denoted as {\small\texttt{conv\_1}}, {\small\texttt{conv\_2}}, {\small\texttt{conv\_3}}, {\small\texttt{conv\_4}}, and {\small\texttt{conv\_5}}).
For the MobileNet-V2, we selected the output feature of the 4, 7, 11, 14, 18 layers as the raw regional feature (denoted as {\small\texttt{layer\_4}}, {\small\texttt{layer\_7}}, {\small\texttt{layer\_11}}, {\small\texttt{layer\_14}}, and {\small\texttt{layer\_18}}).
For fair comparisons, we downsampled feature maps in different layers of each DNN to the height and width of the output feature at the last convolutional layer.
\emph{E.g.}, for intermediate features in VGG-16, feature maps at different layers were downsampled to the size of {\small$14\times14$}.
This makes the learned regional feature {\small$h^{(r)}$} fairly represent the relative discrimination power of each region, thus enabling fair comparisons between regional features through different layers.
All our experiments were run using PyTorch 1.7.1 on Ubuntu 18.04, with the Intel(R) Xeon(R) CPU E5-2637 v4 @ 3.50GHz and one NVIDIA(R) GeForce(R) RTX 2080 Ti GPU.

\end{document}